\documentclass{article}
\usepackage{times}

\usepackage{amsmath,amsfonts,bm}

\def\eqref#1{equation~\ref{#1}}
\def\1{\bm{1}}

\DeclareMathAlphabet{\mathsfit}{\encodingdefault}{\sfdefault}{m}{sl}
\SetMathAlphabet{\mathsfit}{bold}{\encodingdefault}{\sfdefault}{bx}{n}

\usepackage{url}
\usepackage{natbib}

\usepackage{graphicx}
\usepackage{xcolor}
\usepackage{amssymb}
\usepackage{booktabs}
\usepackage{longtable}
\usepackage{array}
\usepackage{fvextra}
\usepackage{multirow}
\usepackage{float}
\usepackage{placeins}
\usepackage[hypertexnames=false,pdftitle={Pixels or Code? Benchmarking Instruction-Guided Visual Artifact Operation}]{hyperref}

\title{Pixels or Code? Benchmarking Instruction-Guided Visual Artifact Operation}

\author{Shun Guo\textsuperscript{1}, Shuizhou Shi\textsuperscript{1}, Boao Wang\textsuperscript{1},\\
Wei Xiong\textsuperscript{1}, Siyu Sun\textsuperscript{2}, Jiangtong Li\textsuperscript{1}, Jie Li\textsuperscript{1}\\[0.5em]
\textsuperscript{1}Tongji University \qquad \textsuperscript{2}Shanghai Jiao Tong University}
\date{}

\begin{document}

\maketitle

\begin{abstract}
Visual artifacts such as charts, technical diagrams, webpages, and document pages underpin scientific, commercial, and engineering workflows.
Editing these artifacts requires direct modification, dependency-aware updates, preservation of unaffected content, and compliance with domain constraints.
We define this capability as Visual Artifact Operation (VAO).
Prior benchmarks test parts of this problem, including cascading updates in charts.
We study direct edits, dependent updates, and preservation across visual domains and pixel, code, and agent paradigms under one image-only protocol.
We introduce Visual Artifact Operation Bench (VAO-Bench), comprising 1,573 manually screened tasks across nine artifact categories and three difficulty levels.
VAO-Bench uses executable artifacts and deterministic rendering to trace reference transformations and annotate direct edits, dependent updates, and preservation constraints.
All systems receive the same source image and instruction; we evaluate pixel-space editors, code-generating multimodal systems, and tool-using agents on direct-edit accuracy, dependency-aware update, content preservation, domain validity, and visual quality.
Experiments show that executable-code systems better support dependent updates, while pixel-space editors offer faster execution and strong visual quality.
Explicitly stating dependencies improves both paradigms, identifying implicit-update inference as a central bottleneck.
VAO-Bench provides a reproducible foundation for dependable visual artifact editing across generation paradigms.
\end{abstract}

\section{Introduction}

Visual generation models have rapidly improved at synthesizing and editing natural images from human instructions~\citep{rombach2022latent,saharia2022photorealistic}.
Many materials used in scientific, commercial, and engineering workflows, however, are visual artifacts such as charts, technical diagrams, webpages, and document pages.
These artifacts are typically authored in executable representations, including HTML, DOT, LaTeX/TikZ, and Python, and rendered deterministically~\citep{zhuo2026factuality}.
Their visible content encodes named components, spatial structures, semantic relationships, and domain-specific constraints.
Editing a visual artifact therefore requires more than producing a visually plausible output: it requires maintaining consistency among coupled elements.

Editing a visual artifact is often specified by a local instruction, but correct execution may require coordinated changes across the artifact~\citep{zhuo2026factuality,li2026charts}.
For example, inserting a decision node into a flowchart requires reconnecting the relevant paths.
Changing a chart's alert threshold can alter displayed states, summary counts, and visual encodings.
A successful edit must satisfy three requirements: execute the requested direct edit, update every visible element that depends on it, and preserve unrelated content, layout, and domain constraints.
We call this problem \emph{Visual Artifact Operation} (VAO).
Dependency-aware update is the central challenge in VAO because a short, local instruction can require coordinated modifications across multiple regions of an artifact.

\begin{figure}[t]
\centering
\begin{minipage}[c]{0.45\textwidth}
  \centering
   \includegraphics[width=\linewidth]{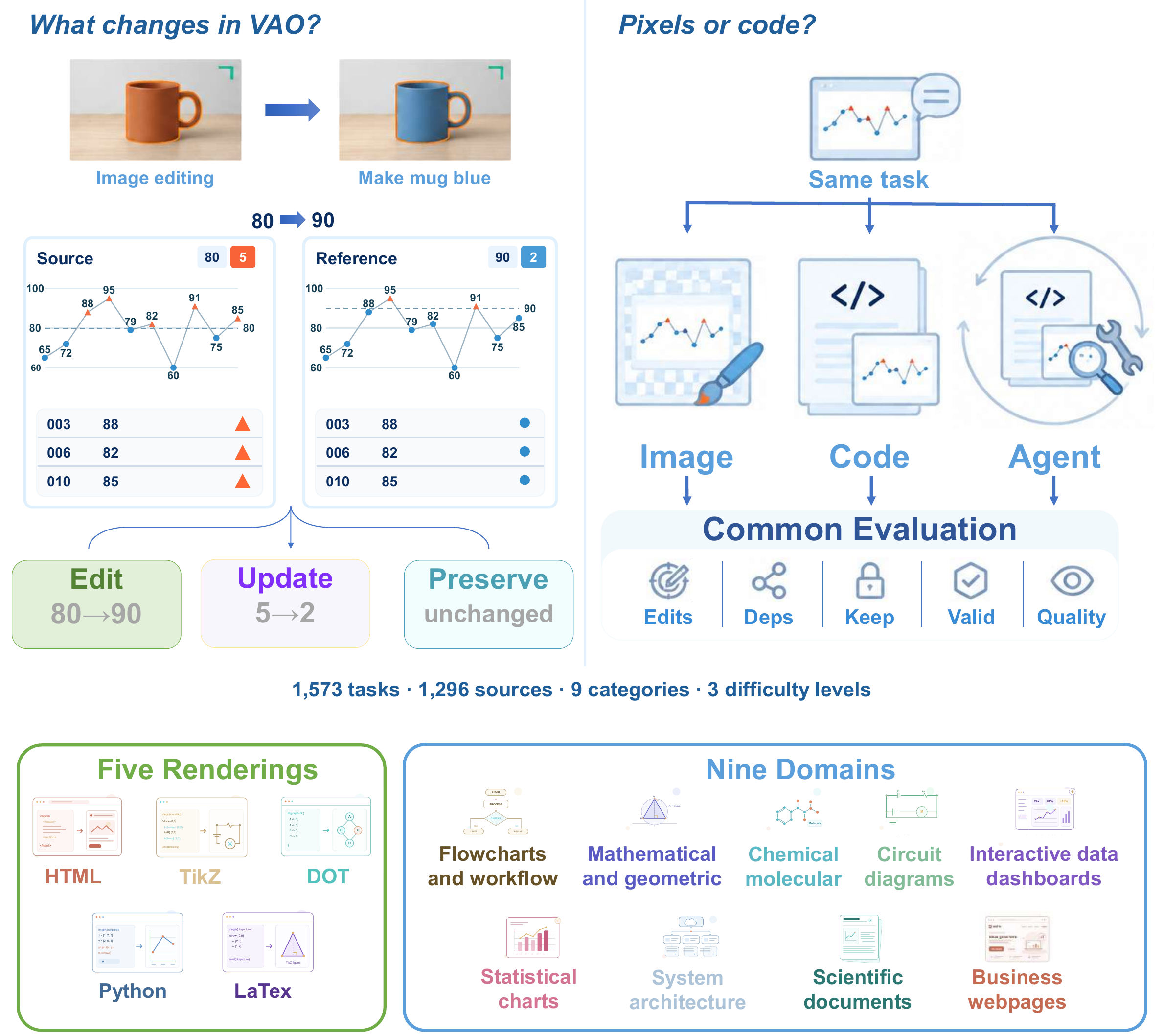}
\end{minipage}%
\hfill%
\begin{minipage}[c]{0.51\textwidth}
  \centering
  \newcommand{\figyes}{\textcolor[HTML]{00969D}{$\checkmark$}}
  \newcommand{\figpartial}{\textcolor[HTML]{6D7885}{$\odot$}}
  \newcommand{\figno}{\textcolor[HTML]{A0AAB3}{$\times$}}
  \resizebox{\linewidth}{!}{%
    \begingroup
    \fontsize{7}{8.6}\selectfont
    \setlength{\tabcolsep}{1.4pt}
    \renewcommand{\arraystretch}{1.06}
    \begin{tabular}{@{}l r c c c c c c@{}}
      \hline
       & \multicolumn{2}{c}{\textbf{Profile}}
      & \multicolumn{3}{c}{\textbf{Domains}}
      & \multicolumn{2}{c}{\textbf{Capabilities}} \\
      \cline{2-3}\cline{4-6}\cline{7-8}
      & Size
      & Code
      & \textbf{Sci.}
      & \textbf{UI}
      & \textbf{Diag.}
      & \textbf{Diff.}
      & \textbf{Deps.} \\
      \hline
      EditBench       & 240   & 0 & \figno  & \figno  & \figno & \figno      & \figno \\
      MagicBrush      & 1,053 & 0 & \figno  & \figno  & \figno & \figno      & \figno \\
      ImgEdit-Bench   & 811   & 0 & \figno  & \figno  & \figno & \figpartial & \figno \\
      \hline
      ChartEditBench  & 5,000 & 1 & \figyes     & \figno  & \figno & \figyes & \figpartial \\
      ChartEdit       & 1,405 & 1 & \figyes     & \figno  & \figno & \figno  & \figno \\
      ChartSync       & 870   & 0 & \figyes     & \figno  & \figno & \figpartial & \figyes \\
      VisEditBench    & 1,395 & 2 & \figpartial & \figno  & \figno & \figyes & \figpartial \\
      DesignBench (Edit) & 359 & 4 & \figno      & \figyes & \figno & \figyes & \figno \\
      \hline
      SVGEditBench V2 & 1,683 & 1 & \figno  & \figno & \figno  & \figno      & \figno \\
      SciDiagramEdit  & 364   & 1 & \figyes & \figno & \figyes & \figno      & \figpartial \\
      Edit2TikZ       & 1,548 & 1 & \figyes & \figno & \figyes & \figpartial & \figpartial \\
      DaEdiTikZ-Bench & 690   & 1 & \figyes & \figno & \figyes & \figno      & \figno \\
      \hline
      \textbf{Ours} & 1,573 & 4 & \figyes & \figyes & \figyes & \figyes & \figyes \\
      \hline
    \end{tabular}
    \endgroup
  }
\end{minipage}%
\par\smallskip
\caption{Visual artifact operation (left) and benchmark comparison (right). Size follows each source's unit. Code counts output code environments, not source-code access. Sci., scientific; UI, interfaces; Diag., diagrams; Diff., difficulty levels or task-complexity analysis; Deps., dependent-update evaluation ($\checkmark$: explicit; $\odot$: related checks; $\times$: absent). Best viewed by zoom in.}
\label{fig:fig1}
\vspace{-4pt}
\end{figure}

Recent benchmarks have expanded evaluation of instruction-guided image editing and visual code generation~\citep{wang2023editbench,si2025design2code}.
Image-editing benchmarks, including EditBench, MagicBrush, and PIE-Bench, assess whether a requested edit is realized and whether source content is visually preserved~\citep{wang2023editbench,zhang2023magicbrush,ju2024pnp}.
Vision-to-code benchmarks, such as Design2Code and ChartMimic, assess reconstruction of webpages or charts from visual references~\citep{si2025design2code,yang2025chartmimic}.
ChartSync explicitly evaluates cascading value-to-geometry updates in charts, comparing image editors with a code-mediated pipeline~\citep{yu2026chartsync}.
These benchmarks establish important capabilities; VAO-Bench combines multiple visual domains with separate evaluation of direct edits, dependent updates, and preservation under one image-only protocol for pixel editors, code generators, and tool-using agents.
Figure~\ref{fig:fig1}~(left) summarizes this capability gap and the evaluation space introduced by VAO.

To evaluate VAO under controlled conditions, we introduce \emph{Visual Artifact Operation Bench}~(\textbf{VAO-Bench}), which constructs tasks from executable artifacts and deterministic renderers.
The pipeline generates an artifact specification and source program, renders the source image, and derives an instruction annotated with direct edits, dependent updates, and preservation constraints.
It then applies the corresponding program-level transformation and deterministically renders a reference target, tracing every required visual consequence to an executable edit.
Multimodal verification and targeted repair ensure that retained artifacts are readable, instructions are visually grounded, and source-target pairs satisfy their annotations.
Solvers receive only the source image and instruction; programs, annotations, and reference targets are retained for benchmark construction and evaluation.

VAO-Bench contains 1,573 manually screened samples spanning nine artifact categories: business webpages, scientific documents, interactive data dashboards, flowcharts and workflow diagrams, system architecture diagrams, circuit diagrams, statistical charts, mathematical and geometric figures, and chemical molecular diagrams.
The collection covers HTML, DOT, LaTeX/TikZ, and Python, enabling dependency-aware editing across visual languages.
Easy, Medium, and Hard tasks progressively test local edits, linked updates, and coordinated modifications under structural or domain constraints.
Figure~\ref{fig:fig1}~(right) compares benchmark profiles, domains, and capabilities.

VAO-Bench supports a unified image-only evaluation of pixel-space editors, multimodal systems that synthesize executable code, and tool-using multimodal agents.
Every system receives the same source image and instruction.
The evaluation measures direct-edit accuracy, dependency-aware updates, content preservation, domain validity, and visual quality.
Tool-using agents perform best overall: Codex-GPT-5.6-Sol leads on requested edits, dependent updates, preservation, and domain validity, while GPT-Image-2 is fastest and achieves the highest visual quality.
Across methods, dependent updates remain harder than requested edits.
Making dependencies explicit yields larger gains than providing the original source code, identifying coupling inference as the central VAO bottleneck.
Our main contributions are:
\begingroup
    \setlength{\topsep}{0pt}
    \setlength{\partopsep}{0pt}
    \setlength{\itemsep}{0pt}
    \setlength{\parsep}{0pt}
\begin{itemize}
    \item We formulate VAO as instruction-guided visual artifact editing that requires direct modification, dependency-aware update, and preservation of unaffected content.
    \item We introduce VAO-Bench, a manually screened benchmark of 1,573 tasks spanning nine artifact categories, four executable representations, and three difficulty levels.
    \item We provide a unified image-only evaluation protocol for comparing pixel generation, executable-code generation, and tool-using agents on VAO.
\end{itemize}
\endgroup

\section{Related Work}

\noindent\textbf{Instruction-Guided Image Editing and Evaluation}\par
Instruction-guided image editing~\citep{ye2025imgedit,zhao2025chartedit} modifies an input image according to a natural-language instruction while preserving content outside the intended change.
Benchmarks such as EditBench, MagicBrush, and PIE-Bench evaluate instruction compliance, edit localization, perceptual quality, and source-content preservation across local and multi-turn editing settings~\citep{wang2023editbench,zhang2023magicbrush,ju2024pnp}.
Recent benchmarks broaden this evaluation to diverse editing operations, complex instructions, LMM-based evaluation, and the balance between modification and preservation~\citep{ma2024i2ebench,xu2025lmm4edit,jia2026compbench,li2025bpm}.
ComplexBench-Edit further evaluates chain-dependent editing instructions in natural images~\citep{wang2025complexbenchedit}.
VAO examines a different dependency pattern: a single local instruction can imply coordinated updates to multiple elements within a structured artifact.
Its evaluation therefore separates the explicitly requested edit, its dependent updates, and the preservation of unaffected content.

\noindent\textbf{Structured Visual Artifact Editing and Reasoning}\par
Structured visual artifacts~\citep{rahman2026viseditbench,xiao2025designbench,nishina2025svgeditbench,sun2026scidiagramedit,zhang2026edit2tikz} encode components, text, spatial organization, data relations, and domain rules, tying semantic correctness to structure.
FigEdit frames scientific chart editing as a structure-aware transformation and evaluates progressively complex edits across diverse chart types~\citep{li2026charts}.
\emph{Factuality Matters} introduces StructBench, built from code-aligned structured visual data, to assess fine-grained factual accuracy in generation and editing across charts, diagrams, and mathematical figures~\citep{zhuo2026factuality}.
GRADE evaluates discipline-informed image editing through Discipline Reasoning, Visual Consistency, and Logical Readability~\citep{liu2026grade}.
ChartSync tests numerical-to-geometric chart updates using a dedicated cascading-update score for 235 cases~\citep{yu2026chartsync}.
ChartEditBench evaluates difficulty-controlled, multi-turn chart-code editing and retention of earlier edits; unlike VAO-Bench, its model input includes the existing code~\citep{kapadnis2026charteditbench}.
These works establish structural fidelity and domain reasoning as central requirements for editing structured visual content.
VAO evaluates requested edits, dependent updates, and preservation separately across nine artifact categories and three system paradigms.

\noindent\textbf{Visual-to-Code and Executable Representations}\par
Executable representations~\citep{greisinger2026editikz} make components and rendering procedures explicit, motivating benchmarks that reconstruct artifacts as code from visual references.
Design2Code evaluates screenshot-to-code webpage implementation, while ChartMimic and Plot2Code assess code generation for charts and scientific plots~\citep{si2025design2code,yang2025chartmimic,wu2025plot2code}.
These benchmarks measure whether multimodal models can recover a renderable representation that reproduces observed visual content.
WebGen-Agent demonstrates agentic website generation through iterative code refinement using screenshot feedback and GUI-agent testing~\citep{lu2026webgenagent}.
VAO instead evaluates instruction-guided transformation of an existing artifact.
Under image-only input, code-based systems must recover a representation from the source image and apply the edit without seeing a reference visual code.
Existing studies evaluate reconstruction or agentic generation in separate settings rather than comparing pixel-space editors, code-generating systems, and tool-using agents under the image-only editing protocol.
VAO provides this comparison to isolate how output representation and iterative tool use support requested edits, dependent updates, and preservation.

\section{Benchmark Construction}
\label{construction}

\begin{figure}[ht]
\begin{center}
\includegraphics[width=\textwidth]{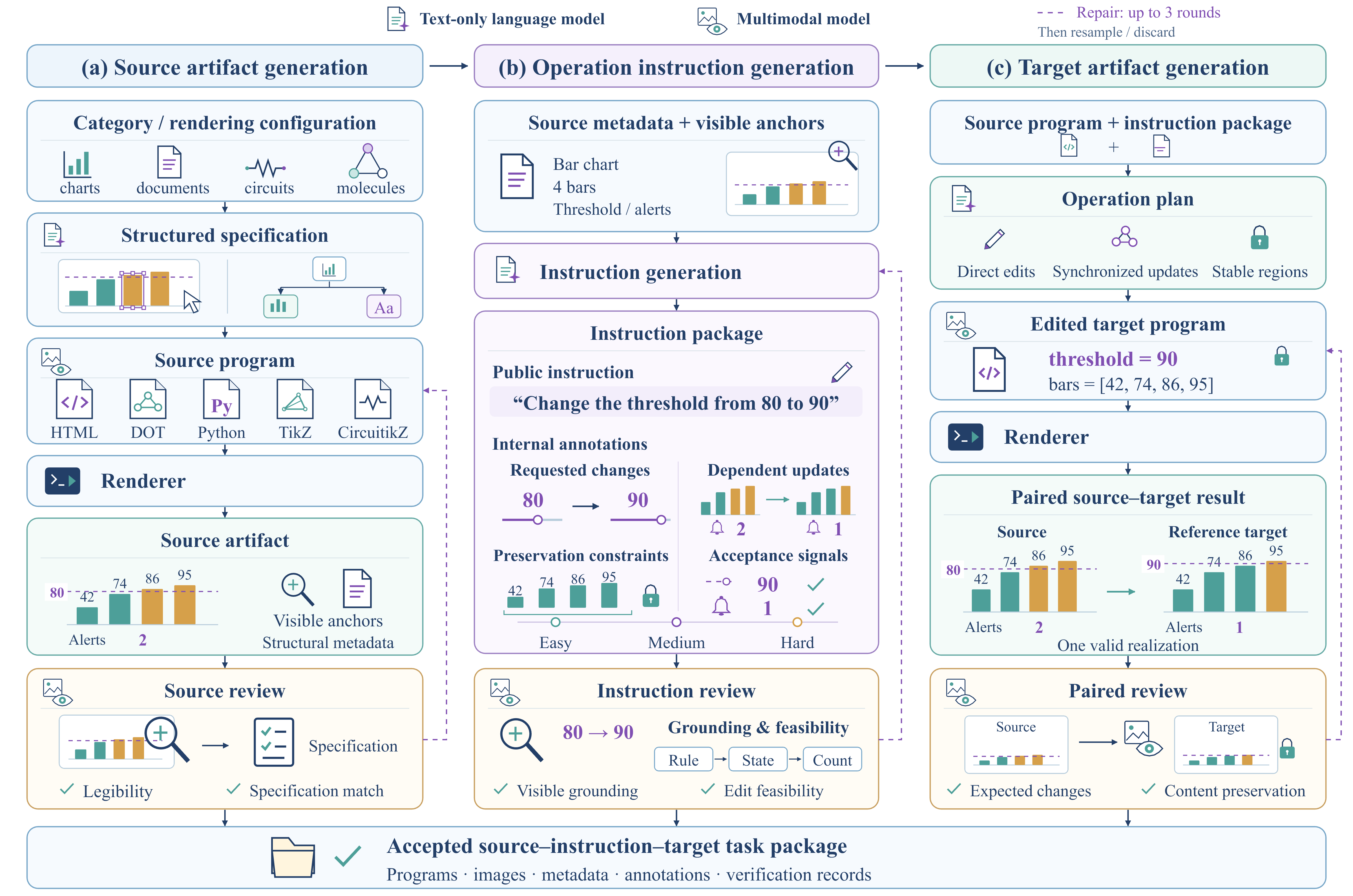}
\end{center}
\caption{Three-stage VAO-Bench construction pipeline: the language model generates specifications, instructions, annotations, and plans; the multimodal model generates and edits visual code and reviews its renderings. Failed checks allow up to three repair rounds. Accepted packages retain code, images, metadata, annotations, and verification records.}
\label{fig:construction_pipeline}
\end{figure}

We construct VAO-Bench from executable visual artifacts using the three-stage pipeline in Figure~\ref{fig:construction_pipeline}.
%We use DeepSeek-V4-Flash as the text-only model and Qwen3.5-397B-A17B as the multimodal model throughout construction.
We use DeepSeek-V4-Flash~\citep{deepseek2026v4} as the text-only model and Qwen3.5-397B-A17B~\citep{qwen2026qwen35} as the multimodal model throughout construction.
In stage (a), the language model generates a structured specification, and the multimodal model generates visual code that a category-specific renderer executes to produce the source image.
In stage (b), the language model generates the public instruction and internal annotations, while the multimodal model reviews the instruction against the source image.
In stage (c), the language model creates an operation plan, and the multimodal model edits the source visual code and reviews the rendered target.
A failed check can trigger up to three repair rounds, after which the candidate is resampled or discarded before human screening.
The resulting task package retains the programs, images, annotations, and verification records; evaluated systems receive only the source image and public instruction.

\subsection{Source Artifact Generation}

This stage begins with an artifact category and rendering configuration, such as an HTML page rendered in a browser or a DOT diagram rendered with Graphviz.
For each, the language model generates a structured specification of visible objects, their layout, and data or semantic relationships among them.
The specification identifies candidate targets, related elements, and content to preserve, so the source can support direct edits and coordinated updates.
Because evaluated systems receive only the source image, evidence needed to identify an operation and infer its consequences must be visually recoverable.
For example, a chart can expose values through labels or an accompanying data table, while a diagram can show relationships through labeled connections.

The multimodal model translates the specification into executable visual code, which the category-specific renderer executes to produce the source image.
Generation instructions constrain visual density and leave room for anticipated operations, such as inserting a component or expanding a label.
The multimodal model reviews the rendered source image against the specification, checking readability, consistency of visible objects and relationships, and whether the intended operation targets are visually supported.
If the source fails review, feedback guides code revision for up to three repair rounds.
Candidates that still fail are discarded or resampled before instruction generation.
An accepted source consists of its executable program and rendered image, together with structural metadata that records visible anchors and relationships passed to the next stage.
The source-generation prompts, metadata fields, and renderer configurations are detailed in App.~\ref{app:source-generation}.

\subsection{Operation Instruction Generation}

Before instruction generation, each candidate task is assigned an operation type and difficulty level according to a predefined stratified schedule that sets candidate quotas for each feasible combination of artifact category, operation type, and difficulty level (App.~\ref{app:instruction-generation}).
The language model receives the visible anchors and relationships from the accepted source metadata as a textual record, then selects an operation target compatible with the assigned settings.
Using visible evidence for that target, it drafts a concise public instruction and corresponding internal annotations.
The instruction identifies its target through evidence such as a label, value, or component name.
The annotations separately record the direct change, dependent updates, and content to preserve.
Keeping dependent updates out of the public instruction allows the benchmark to test whether systems infer these consequences from the source image.
We retain only consequences supported by the instruction and visible source evidence, and keep the annotations out of the evaluated input.

Figure~\ref{fig:construction_pipeline} illustrates the instruction “Change the threshold from 80 to 90” for a dashboard with measurements of 42, 74, 86, and 95 and an alert count of two.
According to the displayed threshold encoding, the measurement at 86 changes from warning to normal, and the alert count decreases from two to one.
The measurement values and record identities remain unchanged.
The internal annotations capture the state change, updated alert count, and preserved values, while the public instruction names only the threshold change.
This example shows how a concise instruction can require coordinated updates across an artifact.

Difficulty levels are defined by the scope of coordination each instruction requires.
Easy tasks primarily change a local label, visual attribute, or state.
Medium tasks require a local edit and linked updates, such as synchronizing a value with its graphical encoding and summary.
Hard tasks alter a rule, threshold, structure, or domain relation, requiring broader propagation or recomputation.
The multimodal model reviews candidate instructions and annotations against the source image.
It checks whether the target is clearly grounded and feasible, whether the annotated consequences are supported by visible evidence, and whether the assigned difficulty matches the required coordination.
Verifier feedback guides up to three rounds of revision or resampling within the requested operation type and difficulty; candidates that still fail review are discarded.
The task schedule, operation taxonomy, difficulty rubric, and annotation schema are provided in App.~\ref{app:instruction-generation}.

\subsection{Target Artifact Generation and Verification}

For each accepted task, the language model uses the instruction, source program, and internal annotations to create an operation plan.
The plan maps the requested direct change, dependent updates, and content to preserve to program-level edits.
The multimodal model applies this plan to the accepted source visual code and produces an edited target program.
The corresponding renderer executes the target program under the source configuration to produce the reference target.
For numerical operations, the plan recomputes dependent values and updates their visible encodings.
For structural operations, it modifies the relevant components and connections.
Deriving the target program from the accepted source links the planned changes to their program-level implementation.

The multimodal model reviews the rendered source-target pair against the public instruction and internal annotations.
It checks that the direct edit and annotated dependent updates are present and mutually consistent, that content marked for preservation remains unchanged, and that the target remains readable and satisfies applicable domain constraints.
Rendering failures or failed visual checks trigger feedback for target-code repair, with at most three repair rounds.
Candidates that still fail review are discarded or resampled.
Each accepted pair provides a validated reference realization, and its source and target programs, images, and verification records are retained for inspection and rerendering.
The full target verification rubric and repair protocol are provided in App.~\ref{app:target-verification}.

After automated verification, human reviewers screen each candidate task package, including the source image, public instruction, target image, and internal annotations.
They check whether the instruction is unambiguous and visually grounded, the scoring rubric is clear and complete, the target realizes the direct edit and dependent updates, preserved content remains intact, and both the source image and target are readable and valid for their artifact domain. 
Candidates that fail these checks are discarded.
The screening checklist and adjudication procedure are provided in App.~\ref{app:human-screening}.

\subsection{Dataset Statistics}

\begin{figure}[t]
\begin{center}
\includegraphics[width=0.9\textwidth]{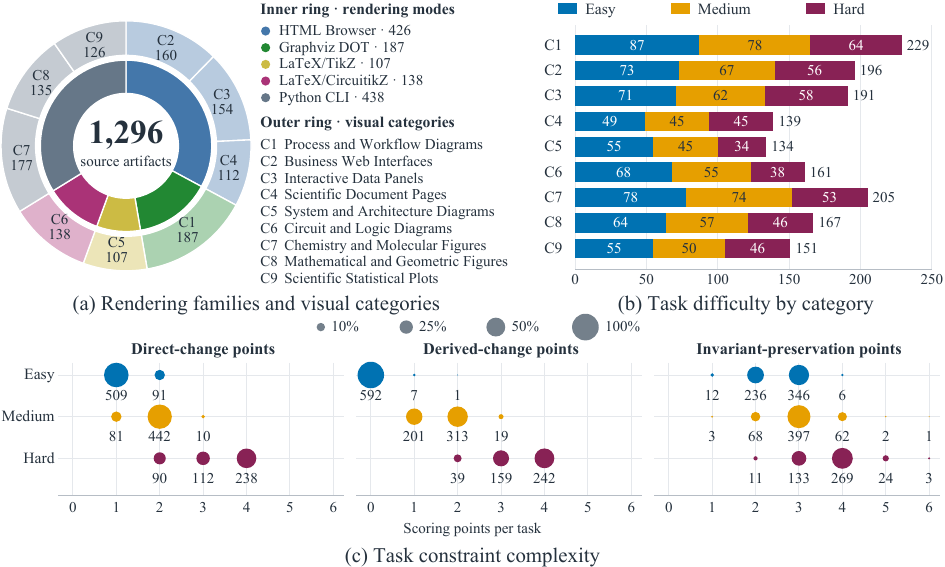}
\end{center}
\caption{Dataset composition and task complexity of VAO-Bench.
The panels summarize artifact-category and rendering-family counts, difficulty by category, and per-task score points for direct changes, dependent updates, and content preservation.}
\label{fig:dataset_statistics}
\end{figure}

After automated construction and human screening, VAO-Bench contains 1,573 tasks from 1296 source artifacts.
It spans nine categories: business webpages, scientific document pages, interactive dashboards, flowchart and workflow diagrams, system architecture diagrams, circuit diagrams, statistical charts, mathematical and geometric figures, and chemical molecular diagrams.
They use four executable representation families: HTML, Graphviz DOT, LaTeX/TikZ (including CircuitikZ), and Python.
The tasks cover 14 operation types and three difficulty levels: Easy, Medium, and Hard.
Figure~\ref{fig:dataset_statistics} will summarize category and rendering-family counts, difficulty by category, and per-task score points for direct changes, dependent updates, and content preservation.

% After automated construction and human screening, Visual Artifact Operation Bench contains 6,781 operation tasks derived from 2,808 source artifacts. The collection comprises 2,592 Easy, 2,292 Medium, and 1,897 Hard tasks. Screening can retain all or only some tasks associated with a source, so a source need not contribute all three difficulty levels. Each retained task pairs an instruction with its original source and corresponding reference target; tasks sharing a source represent separate editing requests.

% The benchmark spans nine artifact categories. HTML supports business web interfaces, scientific document pages, and interactive data panels. Graphviz DOT represents process and workflow diagrams, while LaTeX/TikZ and CircuitikZ represent system architecture and circuit diagrams, respectively. Python generates statistical plots, mathematical and geometric figures, and chemical molecular diagrams. Across these categories, 14 operation types cover local text and appearance changes, linked information updates, structural edits, and operations involving domain constraints. Each task retains its programs, images, annotations, and construction records in a frozen archive. The source image and concise instruction define the editing request, while reference targets and internal annotations are kept separate from solver inputs. This preserves the distinction between constructing a task and evaluating systems on it.

\section{Experiments}
\label{sec:experiments}

% \subsection{Experimental Setup}

\textbf{Models and execution.} 
We evaluate ten methods across three tracks on VAO-Bench. 
%The direct track includes GPT-Image-2 and Seedream-5.0-Pro. 
%The native track includes GPT-5.6-Sol, DeepSeek-Flash, and GLM-5.3-Flash. The agent track pairs Codex with GPT-5.6-Sol, DeepSeek-Flash, or GLM-5.3-Flash, and Claude Code (CC) with DeepSeek-Flash or GLM-5.3-Flash. 
The direct track includes GPT-Image-2~\citep{openai2026gptimage2} and Seedream-5.0-Pro~\citep{bytedance2026seedream}.
The native track includes GPT-5.6-Sol~\citep{openai2026gpt56}, DeepSeek-Flash~\citep{deepseek2026v41flash}, and GLM-5.3-Flash~\citep{zhipu2026glm53flash}.
The agent track pairs Codex~\citep{openai2025codexcli} with GPT-5.6-Sol, DeepSeek-Flash, or GLM-5.3-Flash, and Claude Code (CC)~\citep{anthropic2026claudecode} with DeepSeek-Flash or GLM-5.3-Flash.
Each method receives a source image and a concise instruction, with an execution budget of 1800 seconds per task. 
Direct methods generate images, whereas native and agent methods generate code that is rendered by the evaluation system. 
All methods are evaluated on their final images. 

\textbf{Evaluation.} 
We use Qwen3.5-397B-A17B to assess outputs against the source image, instruction, reference target, and annotated requirements under a common rubric. 
Direct, Derived, and Invariant measure requested edits, dependent updates, and unaffected-content preservation, respectively. 
Each point is binary (1 only if fully satisfied); each dimension is the satisfied-point fraction. 
The judge averages 1-5 visual-quality (VQ) and domain-validity (DQ) ratings across tasks. 
Auto-zero covers all registered tasks, assigning zero to outputs that cannot be executed, rendered, or judged; completed-only averages successfully judged outputs. 
Timed-out runs leaving renderable answers are judged and included in both populations but excluded from the within-budget completion rate. 
We report both populations alongside within-budget completion rate and mean execution time over all registered tasks with timing records. 
Unless otherwise specified, discussion uses auto-zero scores. 
\textbf{Model configurations, prompts, and judging rubrics are provided in the appendix.}

\subsection{Main Results}

\begin{table}[t]
\centering
\caption{Results on VAO-Bench. Comp.: within-budget completion rate; mean time averages all registered tasks with timing records, including timeouts. Dir./Der./Inv.: Direct/Derived/Invariant. Bold marks the highest auto-zero score per column; completed-only scores are not ranked because they cover different task subsets.}
\label{tab:exp-main}
\smallskip
\fontsize{8}{9.5}\selectfont
\setlength{\tabcolsep}{4pt}
\renewcommand{\arraystretch}{1.12}
\resizebox{\linewidth}{!}{%
\begin{tabular}{@{}llrr*{10}{r}@{}}
\toprule
\multirow{2}{*}{Track} & \multirow{2}{*}{Method} & \multirow{2}{*}{Time (s)} & \multirow{2}{*}{Comp.} &
\multicolumn{5}{c}{Auto-zero} & \multicolumn{5}{c}{Completed-only} \\
\cmidrule(lr){5-9}\cmidrule(l){10-14}
 & & & & Dir. & Der. & Inv. & VQ & DQ & Dir. & Der. & Inv. & VQ & DQ \\
\midrule
\multirow{2}{*}{Direct} & GPT-Image-2 & 154.2 & 100\% & 0.730 & 0.527 & 0.916 & \textbf{4.47} & 3.79 & 0.730 & 0.527 & 0.916 & 4.47 & 3.79 \\
 & Seedream-5.0-Pro & 226.8 & 94.0\% & 0.756 & 0.622 & 0.890 & 4.20 & 3.81 & 0.804 & 0.661 & 0.949 & 4.47 & 4.05 \\
\midrule
\multirow{3}{*}{Native} & GPT-5.6-Sol & 300 & 97.2\% & 0.806 & 0.664 & 0.925 & 4.06 & 4.10 & 0.831 & 0.683 & 0.951 & 4.17 & 4.22 \\
 & DeepSeek-Flash & 354 & 94.7\% & 0.778 & 0.631 & 0.826 & 3.99 & 3.81 & 0.825 & 0.671 & 0.874 & 4.21 & 4.02 \\
 & GLM-5.3-Flash & 1353 & 50.4\% & 0.415 & 0.322 & 0.458 & 2.24 & 2.13 & 0.848 & 0.664 & 0.919 & 4.45 & 4.22 \\
\midrule
\multirow{5}{*}{Agent} & Codex-GPT-5.6-Sol & 810 & 96.9\% & \textbf{0.825} & \textbf{0.698} & \textbf{0.957} & 4.43 & \textbf{4.28} & 0.832 & 0.703 & 0.967 & 4.48 & 4.33 \\
 & Codex-DeepSeek-Flash & 1080 & 82.6\% & 0.759 & 0.630 & 0.901 & 4.27 & 4.02 & 0.810 & 0.669 & 0.967 & 4.58 & 4.31 \\
 & Codex-GLM-5.3-Flash & 1308 & 65.6\% & 0.770 & 0.634 & 0.855 & 3.76 & 3.75 & 0.807 & 0.661 & 0.900 & 3.96 & 3.96 \\
 & CC-DeepSeek-Flash & 1100 & 80.4\% & 0.766 & 0.635 & 0.912 & 4.30 & 4.01 & 0.810 & 0.669 & 0.968 & 4.57 & 4.27 \\
 & CC-GLM-5.3-Flash & 1782 & 4.58\% & 0.113 & 0.0879 & 0.203 & 1.03 & 0.907 & 0.591 & 0.500 & 0.957 & 4.63 & 4.08 \\
\bottomrule
\end{tabular}
}
\end{table}

\begin{figure}[t]
\centering
\includegraphics[width=\linewidth]{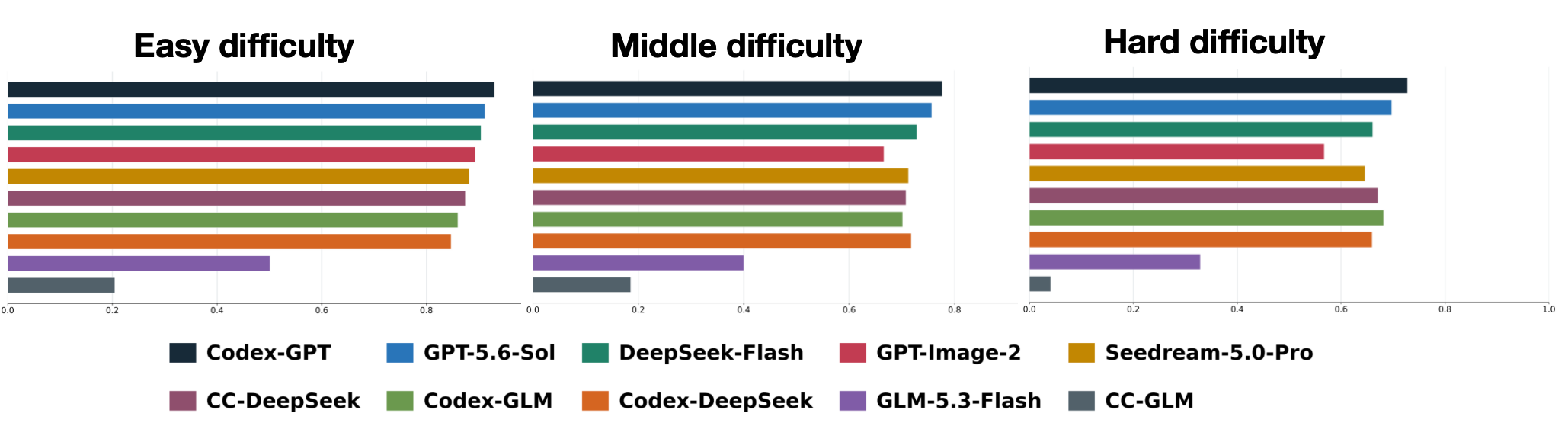}
\caption{Difficulty-stratified auto-zero performance of the ten evaluated systems. Horizontal bars show aggregate requirement scores for Easy, Medium, and Hard tasks, with systems ranked independently within each panel. Execution failures receive zero. \textbf{Best viewed by zoom in}.}
\label{fig:exp-difficulty}
\end{figure}

\textbf{Complementary strengths.} Table~\ref{tab:exp-main} shows that pixel editing and code generation offer different strengths.
GPT-Image-2 is fastest (154.2 seconds) and achieves the highest auto-zero VQ (4.47), but its Derived score is only 0.527.
Native GPT-5.6-Sol raises Derived to 0.664; Codex-GPT-5.6-Sol further improves it to 0.698 and leads on auto-zero Direct, Derived, Invariant, and DQ.
\\\textbf{Agent trade-offs.} Compared with native GPT, Codex-GPT improves Invariant from 0.925 to 0.957 at similar completion (96.9\% \textit{v.s.} 97.2\%), but takes longer (810 \textit{v.s.} 300 seconds).
Both DeepSeek agents outperform native DeepSeek on completed-only Invariant, VQ, and DQ, while Direct and Derived remain similar or slightly lower; their completion rates are lower and execution is substantially longer.
Codex raises GLM's within-budget completion from 50.4\% to 65.6\% and all five auto-zero scores; Claude Code with GLM completes only 4.58\% of tasks within budget.
\\\textbf{Persistent dependency gap.} Across methods, Derived remains below Direct, indicating that inferring dependent edits remains difficult.
In these end-to-end configurations, code-generating systems better support coordinated changes, while agent benefits depend on the model and harness pairing.
\\\textbf{Difficulty trends.} Figure~\ref{fig:exp-difficulty} shows divergence across difficulty levels.
Native and agent systems retain stable rankings from Easy to Hard, indicating consistent performance ordering as tasks become more complex.
In contrast, direct editors progressively fall in rank, suggesting that their strengths weaken when tasks require more coordinated, dependency-aware modifications.

\subsection{Diagnostic Experiments}

We diagnose three factors: implicit-dependency inference in direct editing, reconstruction and source availability in native code generation, and the execution behavior and limitations of agents.
The direct and native ablations use a common 316-task subset with recomputed baselines, follow the main evaluation protocol, and report auto-zero scores.
\par\smallskip\noindent\textbf{Direct: Explicit dependencies reveal an inference bottleneck.}\par
Providing GPT-Image-2 with oracle instructions specifying requested and dependent changes raises Direct from 0.733 to 0.916, Derived from 0.551 to 0.868, and DQ from 3.83 to 4.31, with smaller gains in Invariant and VQ (Table~\ref{tab:exp-direct}).
The instructions are derived from intermediate artifacts in VAO-Bench's target construction; details are provided in the appendix.
These gains show that inferring implicit dependencies, rather than editing alone, is a primary bottleneck for the direct track.
\begin{table}[t]
\centering
\caption{GPT-Image-2 with explicit instructions on the matched 316-task subset.}
\label{tab:exp-direct}
\smallskip
\fontsize{8}{9.5}\selectfont
\setlength{\tabcolsep}{2.5pt}
\renewcommand{\arraystretch}{1.12}
\begin{tabular}{@{}lrr*{10}{r}@{}}
\toprule
\multirow{2}{*}{Condition} & \multirow{2}{*}{Time (s)} & \multirow{2}{*}{Comp.} &
\multicolumn{5}{c}{Auto-zero} & \multicolumn{5}{c}{Completed-only} \\
\cmidrule(lr){4-8}\cmidrule(l){9-13}
 & & & Dir. & Der. & Inv. & VQ & DQ & Dir. & Der. & Inv. & VQ & DQ \\
\midrule
GPT-Image-2 & 157.5 & 100\% & 0.733 & 0.551 & 0.908 & 4.45 & 3.83 & 0.733 & 0.551 & 0.908 & 4.45 & 3.83 \\
+ Explicit instruction & 195.6 & 99.4\% & \textbf{0.916} & \textbf{0.868} & \textbf{0.928} & \textbf{4.45} & \textbf{4.31} & \textbf{0.923} & \textbf{0.876} & \textbf{0.933} & \textbf{4.48} & \textbf{4.34} \\
\bottomrule
\end{tabular}
\end{table}
\begin{table}[t]
\centering
\caption{Native ablations on reconstruction and instruction understanding.}
\label{tab:exp-native}
\smallskip
\fontsize{8}{9.5}\selectfont
\setlength{\tabcolsep}{4pt}
\renewcommand{\arraystretch}{1.12}
\begin{tabular}{lrrrrrr}
\toprule
\multicolumn{7}{c}{A1: image-to-source reconstruction} \\
\midrule
Condition & Completion & SSIM & PSNR & MAE & MSE & SSIM-0 \\
\midrule
DeepSeek-Flash A1 & 92.1\% & 0.796 & 16.4 & 0.0475 & 0.0276 & 0.733 \\
\bottomrule
\end{tabular}

\par\medskip
\setlength{\tabcolsep}{1.5pt}
\renewcommand{\arraystretch}{1.12}
\begin{tabular}{@{}lrr*{10}{r}@{}}
\toprule
\multicolumn{13}{c}{A2/B/C: source availability and instruction explicitness} \\
\midrule
\multirow{2}{*}{Condition} & \multirow{2}{*}{Time (s)} & \multirow{2}{*}{Comp.} &
\multicolumn{5}{c}{Auto-zero} & \multicolumn{5}{c}{Completed-only} \\
\cmidrule(lr){4-8}\cmidrule(l){9-13}
 & & & Dir. & Der. & Inv. & VQ & DQ & Dir. & Der. & Inv. & VQ & DQ \\
\midrule
Baseline & 366 & 95.9\% & 0.798 & 0.652 & 0.842 & 4.02 & 3.88 & 0.826 & 0.677 & 0.879 & 4.19 & 4.04 \\
A2: reconstructed source & 101.4 & 96.2\% & 0.792 & 0.646 & 0.850 & 4.02 & 3.94 & 0.823 & 0.675 & 0.887 & 4.17 & 4.10 \\
B: true source & 87 & 98.4\% & 0.828 & 0.696 & \textbf{0.959} & \textbf{4.63} & \textbf{4.37} & 0.843 & 0.710 & \textbf{0.977} & \textbf{4.70} & \textbf{4.44} \\
C: explicit instruction & 345 & 95.3\% & \textbf{0.919} & \textbf{0.886} & 0.821 & 4.05 & 4.19 & \textbf{0.972} & \textbf{0.938} & 0.867 & 4.26 & 4.40 \\
\bottomrule
\end{tabular}
\end{table}
\\\textbf{Native: Implicit-dependency inference dominates reconstruction.} We evaluate native DeepSeek-Flash under three conditions (Table~\ref{tab:exp-native}): A reconstructs image-derived source code (A1) and edits it (A2), B provides true source code, and C provides oracle explicit instructions.
For A1, we render reconstructed code and compare it with the source image using SSIM, PSNR, MAE, MSE, and SSIM-0; it achieves an SSIM of 0.796.
A2 closely matches the baseline, and reasoning traces confirm the baseline drafts a source-like program before editing it.
True source code improves Invariant, VQ, and DQ but yields only modest Direct and Derived gains.
By contrast, explicit instructions raise Direct and Derived to 0.919 and 0.886, exceeding all main methods on the subset and slightly surpassing explicitly instructed GPT-Image-2, although Invariant decreases.
This contrast identifies implicit-dependency inference as a primary native-track bottleneck, even when accurate source code is available.
B also resembles Codex-GPT on the subset; Codex-GPT traces show iterative rendering and revision toward the source image before target-code editing, suggesting that iterative source reconstruction contributes to its agent advantage alongside tool use.
\begin{table}[t]
\centering
%\caption{Trace statistics. Write, render, and preview columns concern the designated task artifact, excluding local calibration tests. The CC--GLM-5.3-Flash row covers 1,573 inspected episodes, only 72 of which completed successfully.}
\caption{Trace statistics. Write, render, and preview counts refer to the designated task artifact, excluding local calibration tests. Despite averaging 29.4 tool calls per episode, CC--GLM-5.3-Flash completes only 72 of 1,573 episodes, compared with 1,032 for Codex--GLM-5.3-Flash and 1,525 for Codex--GPT-5.6-Sol.}
\label{tab:exp-traces}
\smallskip
\fontsize{8}{10}\selectfont
\setlength{\tabcolsep}{4pt}
\renewcommand{\arraystretch}{1.12}
\begin{tabular}{lrrrrrr}
\toprule
System & Episodes & Completed & \shortstack{Avg. tool\\calls} & \shortstack{Writes\\artifact} & Renders & \shortstack{Views\\preview} \\
\midrule
Codex--GPT-5.6-Sol & 1573 & 1525 & 10.6 & 1562 & 1499 & 1334 \\
Codex--GLM-5.3-Flash & 1573 & 1032 & 25.9 & 1572 & 1550 & 514 \\
CC--GLM-5.3-Flash & 1573 & 72 & 29.4 & 365 & 343 & 103 \\
\bottomrule
\end{tabular}
\end{table}
\\\textbf{Agent: Implicit-dependency inference remains the an important bottleneck.} Under the common 30-minute budget, we compare native DeepSeek with Codex and Claude Code agents in Table~\ref{tab:exp-main} and Figure~\ref{fig:exp-difficulty}.
The agents improve Invariant, VQ, and DQ but require more time, complete fewer tasks, and do not improve Derived.
Along with the explicit-instruction gains in Tables~\ref{tab:exp-direct} and~\ref{tab:exp-native}, these results suggest implicit-dependency inference remains an important bottleneck on VAO.
\\\textbf{Agent: Harness compatibility governs execution.} We compare agent outcomes in Table~\ref{tab:exp-main} and Figure~\ref{fig:exp-difficulty}, and analyze execution traces summarized in Table~\ref{tab:exp-traces}.
DeepSeek agents achieve strong completed-only Invariant, VQ, and DQ, but complete fewer than 83\% of tasks within budget and retain low Direct and Derived scores.
For GLM, Claude Code traces show prolonged analysis before artifact construction, corresponding to only 4.58\% completion, whereas Codex enables earlier drafting and iterative verification, increasing completion by 61.0 percentage points.
GPT-5.6-Sol similarly implements early and then renders and revises, leading agents in speed, completion, and all auto-zero metrics.
Thus, agent performance depends jointly on model capability and harness compatibility; 
%trace-analysis details are provided in the appendix.
trace-analysis details are provided in App.~\ref{app:trace-analysis-and-completion-rate-statistics}.

\subsection{Case Study}

\begin{figure}[t]
\centering
\includegraphics[width=\linewidth]{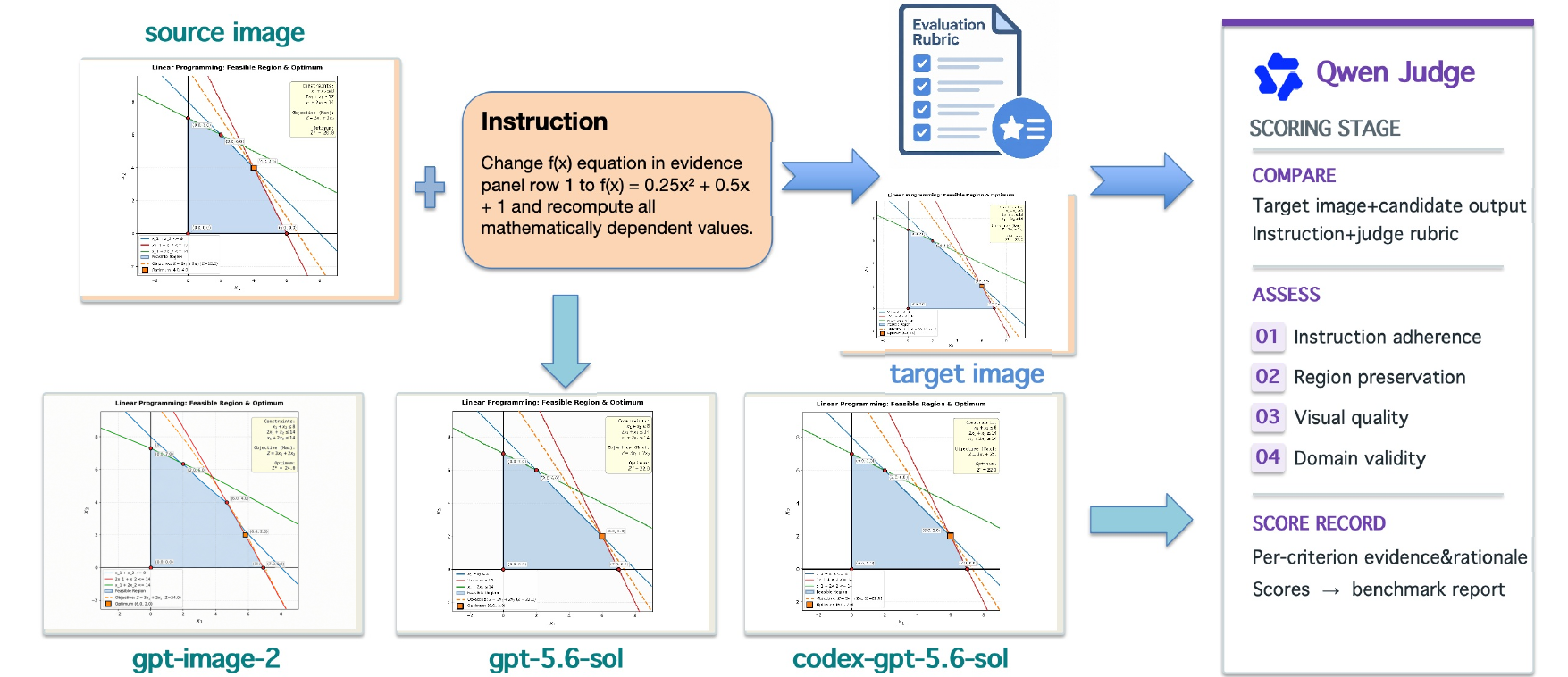}
\caption{Case study of dependency-aware VAO. Changing the objective function requires updating dependent elements while preserving the constraints, region, and layout. \textbf{Best viewed by zoom in.}}
\label{fig:case-study}
\end{figure}

Figure~\ref{fig:case-study} illustrates how a local textual edit can require coordinated changes throughout an artifact.
Changing the objective equation affects its plotted curve and derived annotations, while unrelated constraints, the feasible region, and the visual structure should remain unchanged.
The outputs from GPT-Image-2, native GPT-5.6-Sol, and Codex--GPT-5.6-Sol demonstrate different balances between instruction adherence and preservation, which the judge assesses against the executable reference target under the same image-only protocol using criterion-specific evidence for task accuracy, preservation, visual quality, and domain validity.
% This case shows why visual plausibility alone is insufficient: correctness also requires coherent propagation of the mathematical change.

\section{Conclusion}

We introduced Visual Artifact Operation (VAO), a unified formulation of visual artifact editing that requires requested edits, dependent updates, and preservation of unaffected content.
We also presented VAO-Bench, 1,573 manually screened tasks spanning nine artifact categories, four executable representations, and three difficulty levels, with a unified image-only protocol for pixel editors, code-generating systems, and tool-using agents.
Results reveal complementary strengths: pixel editors are fast and visually strong, whereas executable-code systems better support coordinated updates.
However, dependent updates remain harder than direct edits across paradigms, and explicit dependency instructions help more than access to true source code.
This identifies implicit-dependency inference, rather than reconstruction alone, as the central VAO bottleneck and positions VAO-Bench as a reproducible foundation for developing more dependable visual artifact editors.

% Appendix material (not typeset):
% - Model configurations, execution budgets, prompts, and judging rubrics.
% - Full domain, rendering-mode, and difficulty results, including scoring-point denominators and timeout recovery.
% - Additional reconstruction metrics, explicit-instruction construction, ablation input examples, and stage-specific execution times.
% - Trace sampling and counting procedures, representative revision sequences, and failure cases.

\bibliography{references}
\bibliographystyle{plainnat}

\appendix

%\section{Appendix}
%You may include other additional sections here.

% Appendix definitions and inputs follow.
\definecolor{promptbg}{RGB}{248,249,250}
\definecolor{promptline}{RGB}{190,196,204}
\setlength{\LTpre}{4pt}
\setlength{\LTpost}{4pt}

\section{Dataset Composition and Construction Details}
\label{app:dataset}
\subsection{Dataset Composition and Units of Analysis}
VAO-Bench contains 1,573 tasks based on 1,296 source artifacts. Each task consists of a source image and an editing instruction; tasks sharing a source image are executed independently. The Easy, Medium, and Hard levels contain 600, 533, and 440 tasks, respectively. Four source-code representations support five rendering modes: HTML, DOT, Python, TikZ, and CircuitikZ.

\begingroup\footnotesize
\setlength{\tabcolsep}{3pt}
\begin{longtable}{llrrrrrr}
\caption{Composition of VAO-Bench and its ablation subset, VAO-Bench-Ablation. Source counts refer to distinct source artifacts.}\label{tab:app-composition}\\
\toprule
Category & Renderer & Sources & Easy & Medium & Hard & Main & Subset\\\midrule
\endfirsthead
\toprule
Category & Renderer & Sources & Easy & Medium & Hard & Main & Subset\\\midrule
\endhead
\bottomrule\endfoot
Business webpages & HTML & 160 & 73 & 67 & 56 & 196 & 40\\
Scientific documents & HTML & 112 & 49 & 45 & 45 & 139 & 28\\
Interactive dashboards & HTML & 154 & 71 & 62 & 58 & 191 & 38\\
Flowcharts and workflows & DOT & 187 & 87 & 78 & 64 & 229 & 46\\
Systems and architecture & TikZ & 107 & 55 & 45 & 34 & 134 & 26\\
Circuits and logic & CircuitikZ & 138 & 68 & 55 & 38 & 161 & 33\\
Scientific statistical charts & Python & 126 & 55 & 50 & 46 & 151 & 31\\
Mathematics and geometry & Python & 135 & 64 & 57 & 46 & 167 & 34\\
Chemical diagrams & Python & 177 & 78 & 74 & 53 & 205 & 40\\
Total &  & 1296 & 600 & 533 & 440 & 1573 & 316\\
\end{longtable}\endgroup

VAO-Bench-Ablation is a subset of VAO-Bench used for ablation studies. Task-level sampling stratified by source batch, artifact category, and difficulty yields 316 tasks from 304 source artifacts: 120 Easy, 107 Medium, and 89 Hard tasks. All ablation conditions use the same tasks, with baseline results retrieved from the main experiment by task ID.

Difficulty reflects the scope of coordination required. Easy tasks primarily involve local changes to text, appearance, or state. Medium tasks link local values to graphics, legends, or adjacency relations. Hard tasks require propagating changes through rules, thresholds, structures, or domain relations. Eight Easy tasks contain nine derived scoring points in total; Derived is computed over the actual derived points.

\paragraph{Operation coverage.}
Table~\ref{tab:app-operations} lists the 14 operations, ranging from local edits to dependency propagation and domain-consistency repair.
\begingroup\footnotesize
\setlength{\tabcolsep}{3pt}
\begin{longtable}{p{0.70\linewidth}rr}
\caption{The 14 operations in VAO-Bench and their task counts.}\label{tab:app-operations}\\
\toprule
Operation & Difficulty & Tasks\\\midrule
\endfirsthead
\toprule
Operation & Difficulty & Tasks\\\midrule
\endhead
\bottomrule\endfoot
Rename labels and text & Easy & 200\\
Change colors, styles, markers, or badges & Easy & 128\\
Edit annotations, captions, or legends & Easy & 149\\
Change highlights, states, or selections & Easy & 123\\
Update local values and synchronize visual elements & Medium & 178\\
Insert or delete local components, reconnect, and reflow & Medium & 114\\
Move or reorder panels, nodes, or object groups & Medium & 90\\
Update derived labels, summaries, or references & Medium & 97\\
Synchronize encodings and styles within a local group & Medium & 54\\
Propagate global thresholds, filters, or baselines & Hard & 98\\
Refactor structures with dependencies & Hard & 106\\
Recompute derived metrics, geometry, logic, or chemical relations & Hard & 128\\
Transform layouts across panels or regions & Hard & 54\\
Repair domain consistency & Hard & 54\\
\end{longtable}\endgroup

\subsection{Source Artifact Generation and Visible Evidence}
\label{app:source-generation}
DeepSeek-V4-Flash generates specifications, instructions, internal annotations, and operation plans. Qwen3.5-397B-A17B generates and edits visual code and performs multimodal review. Source specifications describe appearance and the objects, values, and relations recoverable from the image. The corresponding renderer produces a source image from the code. Visual review checks readability, semantic consistency, and suitability for subsequent editing. Failed candidates undergo up to three repair rounds; those that remain unacceptable are discarded or resampled.

\begingroup\small
\begin{longtable}{p{.40\linewidth}p{.53\linewidth}}
\caption{Construction metadata and its uses. These internal fields are not supplied to solvers in the main experiment.}\\
\toprule
Field & Meaning and verification\\\midrule
\endfirsthead
\toprule
Field & Meaning and verification\\\midrule
\endhead
\bottomrule\endfoot
\path{visible_objects}, \path{visible_element_ids} & Visible text, nodes, values, and graphical objects with unique or identifiable visual anchors.\\
\path{dependency_graph}, \path{data_or_logic_relations} & Data, structural, or logical dependencies among visible objects that support linked updates.\\
\path{visual_recoverability_contract} & How labels, tables, or connections provide the evidence needed for an edit.\\
\path{editable_objects}, \path{operation_affordance_matrix} & Candidate objects and their supported operations.\\
\path{stable_regions}, \path{stable_invariants} & Regions, identities, values, or relations to preserve, extending beyond rectangular pixel regions.\\
\end{longtable}\endgroup

Circuit tasks can start from predefined CircuitikZ templates, using component values and permitted editing regions to generate source and target code while retaining template identifiers. Templates constrain structure; review further checks connections, labels, and domain correctness. Rendering settings appear in Appendix~\ref{app:renderers}, and the organization of specification, code, and repair prompts appears in the Supplementary Material.

\subsection{Instruction Scheduling, Difficulty, and Annotations}
\label{app:instruction-generation}

Candidate generation follows a predefined stratified schedule that sets candidate quotas for each feasible combination of artifact category, operation type, and difficulty level. These quotas are assigned before instruction generation and determine the candidate distribution across artifact categories, operation types, and difficulty levels. The schedule is used to organize candidate generation; automated verification and subsequent human screening determine which candidates are retained in the released dataset.

The scheduler first allocates source-artifact slots across artifact categories and then assigns operation types and difficulty levels within each feasible category. Quota weights are converted into integer candidate counts, and the resulting candidate slots are distributed across the feasible combinations specified by the construction configuration. Each source artifact can therefore receive instructions at different difficulty levels, while the assigned operation must remain compatible with the artifact category and subtype. This procedure provides the stratified schedule referenced in the main text.

Using visible anchors in the source metadata, the generator selects objects and produces a concise public instruction together with private requirements for the assigned category, operation type, and difficulty level. The public instruction states the triggering edit, while the private requirements record expected direct changes, dependent updates, and stable invariants. Visual review checks object localization, operation feasibility, visible support for dependencies, and whether the assigned difficulty matches the coordination required by the source artifact. Candidates that fail these checks are revised or discarded before later target-artifact construction.

Internal records contain \path{operation_type}, \path{target_objects}, \path{expected_delta}, \path{required_synchronized_updates}, \path{stable_invariants}, and \path{scoring_points}. Each scoring point includes an ID, type, description, observable region, and expected evidence. Direct covers explicitly requested changes, Derived covers updates implied by dependencies, and Invariant covers content to preserve.

\subsection{Target Verification and Repair}
\label{app:target-verification}
An operation plan specifies objects to modify or preserve, required computations, layout and semantic updates, and risk checks before editing the accepted source code. Paired review examines the source image, target image, public instruction, and internal annotations. It checks direct changes, synchronized dependent updates without stale values, preservation, readability, and domain relations. Failures distinguish rendering errors from renderable but semantically invalid outputs, with feedback used for up to three target repair rounds. Accepted artifacts retain code, images, metadata, and review records. A difference from the source image alone does not establish success.

\subsection{Human Screening and Access Boundaries}
\label{app:human-screening}
The automated pipeline generated 2,098 source images and 6,294 tasks, with one Easy, Medium, and Hard task per image. Five reviewers screened the candidates and discussed disputed tasks collectively, retaining 1,573 tasks.

Screening proceeded in two stages. Reviewers first checked source-image readability and clarity; problems such as overlap or ambiguous presentation led to rejection of all tasks associated with that image. They then reviewed each task for a clear instruction, complete and detailed scoring criteria, and a correct target edit, rejecting any task that failed a check. Main evaluation inputs comprise the source image and instruction; code-based tracks additionally receive the language, rendering target, and allowed dependencies. Source code, reference targets, and scoring requirements are used for construction and evaluation.

\section{Models, Execution, and Evaluation Protocol}
\label{app:protocol}
\subsection{Model Configurations and Inputs}
\label{app:model-settings}
All three tracks receive the same source image and concise instruction, with a main-experiment budget of 1,800 s per task. The direct track returns an image; the native track returns a complete code block; and the agent track writes the designated \texttt{artifact.*} file in the task workspace and can use local tools to render, inspect, and revise it. Solvers receive neither target images nor scoring requirements. Table~\ref{tab:app-models} lists all ten evaluated methods and their recorded settings. Unspecified parameters, including temperature, retain server defaults.

\begingroup\small
\begin{longtable}{lp{.30\linewidth}p{.46\linewidth}}
\caption{Evaluated methods and recorded configurations. M1--M10 identify methods in subsequent tables.}\label{tab:app-models}\\
\toprule
ID & Method / track & Configuration\\\midrule
\endfirsthead
\toprule
ID & Method / track & Configuration\\\midrule
\endhead
\bottomrule\endfoot
M1 & GPT-Image-2 / direct & \texttt{quality=high}; concurrency 5; launch interval 1 s.\\
M2 & Seedream-5.0-Pro / direct & Service ID:\newline {\footnotesize\path{doubao-seedream-5-0-pro-260628}}\newline 1.5K; watermark disabled; concurrency 10; interval 1 s.\\
M3 & GPT-5.6-Sol / native & \texttt{reasoning\_effort=xhigh}; maximum output 65,536 tokens.\\
M4 & DeepSeek-Flash / native & Thinking enabled; \texttt{reasoning\_effort=max}; maximum output 65,536 tokens.\\
M5 & GLM-5.3-Flash / native & \texttt{reasoning\_effort=max}; maximum output 65,536 tokens.\\
M6 & Codex-GPT / agent & Codex + GPT-5.6-Sol; reasoning level medium.\\
M7 & Codex-DeepSeek / agent & Codex + DeepSeek-Flash; reasoning level high.\\
M8 & Codex-GLM / agent & Codex + GLM-5.3-Flash; reasoning level high.\\
M9 & CC-DeepSeek / agent & Claude Code + DeepSeek-Flash; effort high.\\
M10 & CC-GLM / agent & Claude Code + GLM-5.3-Flash; effort high.\\
\end{longtable}\endgroup

The native configuration uses concurrency 10 and a 250 ms request launch interval; rendering concurrency is 4. Infrastructure retries are capped at 3 with a 5 s backoff. The illustrated Codex traces use CLI 0.150.1, and the Claude Code comparison traces use 2.1.250.

\subsection{Canonical Rendering and Tools}
\label{app:renderers}
\begingroup\small
\begin{longtable}{lp{.73\linewidth}}
\caption{Renderers, resolutions, and dependencies used in evaluation.}\\
\toprule
Mode & Settings and constraints\\\midrule
\endfirsthead
\toprule
Mode & Settings and constraints\\\midrule
\endhead
\bottomrule\endfoot
HTML & Headless Chromium; 1440$\times$1000 viewport; full-page screenshot; 60 s rendering timeout; self-contained HTML/CSS/JS with no remote resources.\\
DOT & Graphviz; PNG at 600 DPI; white background; 24 px crop margin; 90 s rendering timeout.\\
TikZ & \texttt{pdflatex} followed by \texttt{pdftoppm} for PNG conversion; 600 DPI; white background; 96 px crop margin; 90 s timeout; tikz, xcolor, and amsmath allowed.\\
CircuitikZ & TikZ settings with circuitikz additionally allowed; recorded as a separate rendering mode.\\
Python & Code runs in the evaluation environment and produces PNG output; allowed packages include Pillow, Matplotlib, NumPy, SciPy, NetworkX, and RDKit.\\
\end{longtable}\endgroup

Agents may generate previews during execution; the evaluator renders the final code again to obtain the official output. The direct track uses the image returned by the model. Model requests, agent sessions, and subsequent canonical rendering are timed separately; aggregation is described in Appendix~\ref{app:populations}.

\subsection{Scoring Points, Visual Quality, and Domain Validity}
\label{app:scoring}
The judge, Qwen3.5-397B-A17B, receives the source image, candidate image, reference target, original instruction, and scoring points. Images are limited to 1,568 px on the longest side. The configuration uses reasoning level xhigh, a maximum output of 64,000 tokens, up to 2 parsing retries and 3 transient-error retries, and a 900 s request limit. Each point receives a binary score, specific visible evidence, a rationale, and a confidence estimate. Points that cannot be verified from the candidate image receive 0. The reference target is one valid implementation; nonessential differences in fonts, antialiasing, or layout are not penalized on their own.

Let $P_{ik}$ denote the scoring points of type $k$ for task $i$ in set $S$, with $s_{ip}\in\{0,1\}$. We compute
\begin{equation}
 A_k(S)=\frac{\sum_{i\in S}\sum_{p\in P_{ik}}s_{ip}}{\sum_{i\in S}|P_{ik}|},
 \qquad k\in\{\mathrm{Direct,Derived,Invariant}\}.
\end{equation}
This aggregation weights scoring points equally. Tasks without points of a given type contribute to neither its numerator nor its denominator; a zero denominator is reported as unavailable. VQ and DQ are averaged separately over tasks.

\begingroup\footnotesize
\setlength{\tabcolsep}{3pt}
\begin{longtable}{lrrrrr}
\caption{Total scoring points / mean per task. The last column counts tasks without Derived points.}\label{tab:app-points}\\
\toprule
Difficulty & Tasks & Direct & Derived & Invariant & No Derived\\\midrule
\endfirsthead
\toprule
Difficulty & Tasks & Direct & Derived & Invariant & No Derived\\\midrule
\endhead
\bottomrule\endfoot
Easy & 600 & 691 / 1.15 & 9 / 0.01 & 1546 / 2.58 & 592\\
Medium & 533 & 995 / 1.87 & 884 / 1.66 & 1594 / 2.99 & 0\\
Hard & 440 & 1468 / 3.34 & 1523 / 3.46 & 1635 / 3.72 & 0\\
\end{longtable}\endgroup

VQ scores of 5/4/3/2/1 indicate clean and complete output, minor blemishes, noticeable but usable defects, severely impaired readability, and largely unusable output, respectively. DQ uses nine category-specific rubrics: state and information consistency for webpages and dashboards; organization and scientific presentation for documents; connections and logic for flowcharts and architecture diagrams; electrical connections for circuits; data encoding for statistical charts; mathematical relations for mathematical figures; and atoms, bonds, and reaction notation for chemical diagrams. The full rubrics are provided in \texttt{prompts/}. Domain scoring does not require pixel-level agreement with the reference.

\subsection{Failures, Timeout Recovery, and Denominators}
\label{app:populations}
For method $m$, let $E_m$ denote registered execution tasks, $J_m$ normally completed and successfully scored tasks, $R_m$ successfully scored tasks recovered after timeout, and $F_m$ the remaining confirmed execution failures. Completed-only scores use $J_m\cup R_m$. Auto-zero covers all registered tasks in $E_m$ and assigns zero to each task in $F_m$ for its scoring points and VQ/DQ. The judge uses a 1--5 scale for valid images.

Normal completion rate divides normally completed tasks by $|E_m|$; completion with recovery additionally counts recovered outputs. Recovered outputs retain their original timeout status. Timing with recovery replaces the corresponding original durations with recovery durations; other tasks use total generation time, falling back to model time when unavailable. Means cover all executed tasks with timing records in the group, including failures.

\begingroup\footnotesize
\setlength{\tabcolsep}{3pt}
\begin{longtable}{lrrrrr}
\caption{Execution coverage. Recovered outputs are a subset of timeouts and are not additional tasks. Other failures include model, code, and rendering errors.}\label{tab:app-coverage}\\
\toprule
Method & Registered & Normal & Recovered & Timeout & Other\\\midrule
\endfirsthead
\toprule
Method & Registered & Normal & Recovered & Timeout & Other\\\midrule
\endhead
\bottomrule\endfoot
GPT-Image-2 & 1573 & 1573 & 0 & 0 & 0\\
Seedream-5.0-Pro & 1573 & 1478 & 0 & 95 & 0\\
GPT-5.6-Sol & 1573 & 1529 & 0 & 44 & 0\\
DeepSeek-Flash & 1573 & 1489 & 0 & 84 & 0\\
GLM-5.3-Flash & 1573 & 793 & 0 & 774 & 6\\
Codex-GPT & 1573 & 1525 & 31 & 33 & 15\\
Codex-DeepSeek & 1573 & 1299 & 168 & 248 & 26\\
Codex-GLM & 1573 & 1032 & 461 & 519 & 22\\
CC-DeepSeek & 1573 & 1265 & 214 & 281 & 27\\
CC-GLM & 1573 & 72 & 277 & 1501 & 0\\
\end{longtable}\endgroup

For execution-coverage reporting, all ten methods are counted over the full 1,573-task benchmark.Table~\ref{tab:app-coverage} reports normal completions, timeouts, and recoveries separately.

\section{Diagnostic Ablations and Agent Behavior}
\label{app:ablations}
\subsection{Explicit Instructions and Controlled Variables}
\label{app:explicit-instructions}
Direct and native ablations use all 316 tasks in VAO-Bench-Ablation. Expanded instructions retain the original text and add content from \path{target_objects}, \path{expected_delta}, and \path{required_synchronized_updates} in construction records, and \path{synchronized_updates_applied} in target metadata. Rule-based extraction and deduplication retain target objects, required changes, and linked updates while filtering preservation requirements, verification steps, and speculative statements.

Expansion changes only solver inputs; evaluation uses the same original instruction, source image, target image, and scoring points. In the direct ablation, Derived increases from 0.5514 to 0.8683, while VQ changes from 4.4494 to 4.4525.

\begin{quote}\begin{minipage}{\linewidth}\small\raggedright
\textbf{Concise versus explicit instructions.} The following summarizes the order-dashboard task \texttt{html\_business\_crm\_ecommerce\_report\_0046} (Hard).

\textbf{Concise:} Change Min Order Threshold from \$50 to \$40 in the filter control.

\textbf{Explicit:} Retain the request above and add: change the risk flags for ORD-2848 (\$43.00) and ORD-2849 (\$45.00) from Yes to No; change revenue at risk from \$126.50 to \$38.50 and the at-risk order count from 3 to 1; update the rule to ``Orders < \$40''; remove risk highlighting from these two rows, retaining it only for ORD-2850.
\end{minipage}\end{quote}

\subsection{Native Reconstruction and Understanding Ablations}
\label{app:native-ablation}
\begingroup\small
\begin{longtable}{lp{.35\linewidth}p{.39\linewidth}}
\caption{Inputs to the DeepSeek-Flash ablations. No editing condition receives the target image.}\\
\toprule
Condition & Inputs & Evaluation and interpretation\\\midrule
\endfirsthead
\toprule
Condition & Inputs & Evaluation and interpretation\\\midrule
\endhead
\bottomrule\endfoot
Baseline & Source image, original instruction & Main-experiment baseline on the same tasks.\\
A1 & Source image; no editing instruction or ground-truth code & Source reconstruction only; offline pixel metrics and rendering success.\\
A2 & Source image, A1 code, original instruction & Edit the reconstructed representation; standard editing evaluation.\\
B & Source image, ground-truth source code, original instruction & Control source-representation availability; standard editing evaluation.\\
C & Source image, expanded instruction; no ground-truth code & Control instruction explicitness; standard editing evaluation.\\
\end{longtable}\endgroup

A1 independently processes all 316 tasks without deduplicating shared source images. It successfully rendered 291 tasks (92.09\%); 14 returned empty responses and 11 encountered compilation or execution errors. Successful outputs had mean SSIM 0.7958, PSNR 16.37 dB, MAE 0.04748, and MSE 0.02757. Assigning zero to failures gives an all-task SSIM of 0.7328. Source images are resized to at most 1,568 px on the longest side; candidates are fitted proportionally to the same canvas and centered with white padding, without stretching, cropping, or registration. SSIM uses scikit-image 0.25.2 with an 11$\times$11 local Gaussian window on RGB channels, $\sigma=1.5$, population covariance, and \texttt{data\_range=1}.

The 14 A1 outputs without code enter A2 as empty-source inputs; the 11 compilation failures retain their original code. A2's 101.4 s measures second-stage editing; total duration is the sum of A1 reconstruction and A2 editing. Baseline, B, and C editing times are 366 s, 87 s, and 345 s, respectively.

\subsection{Agent Execution Strategies and Time-Budget Reliability}
\label{app:trace-analysis-and-completion-rate-statistics}

To further characterize how different agent--model combinations execute VAO tasks, we analyze their execution traces at the level of semantic strategy actions.
We define a set of 15 strategy actions that represent the major behaviors involved in solving a VAO task.
These actions include initial task understanding (viewing the original image, reading \texttt{TASK.md}, and identifying modification scope and preservation requirements), visual reverse engineering (cropping regions, scanning RGB boundaries, measuring text bounding boxes, and analyzing layout or connections), reconstruction and tuning (finding fonts, checking rendering capabilities, and testing graphics parameters), rendering and comparison (rendering previews, performing visual comparison, and comparing pixels), and answer review and editing (reviewing and editing the final answer).
This abstraction groups low-level tool events according to their functional role in the task-solving process and therefore provides a more interpretable representation of agent execution strategy.

\begin{figure}[t]
\centering
\includegraphics[width=\linewidth]{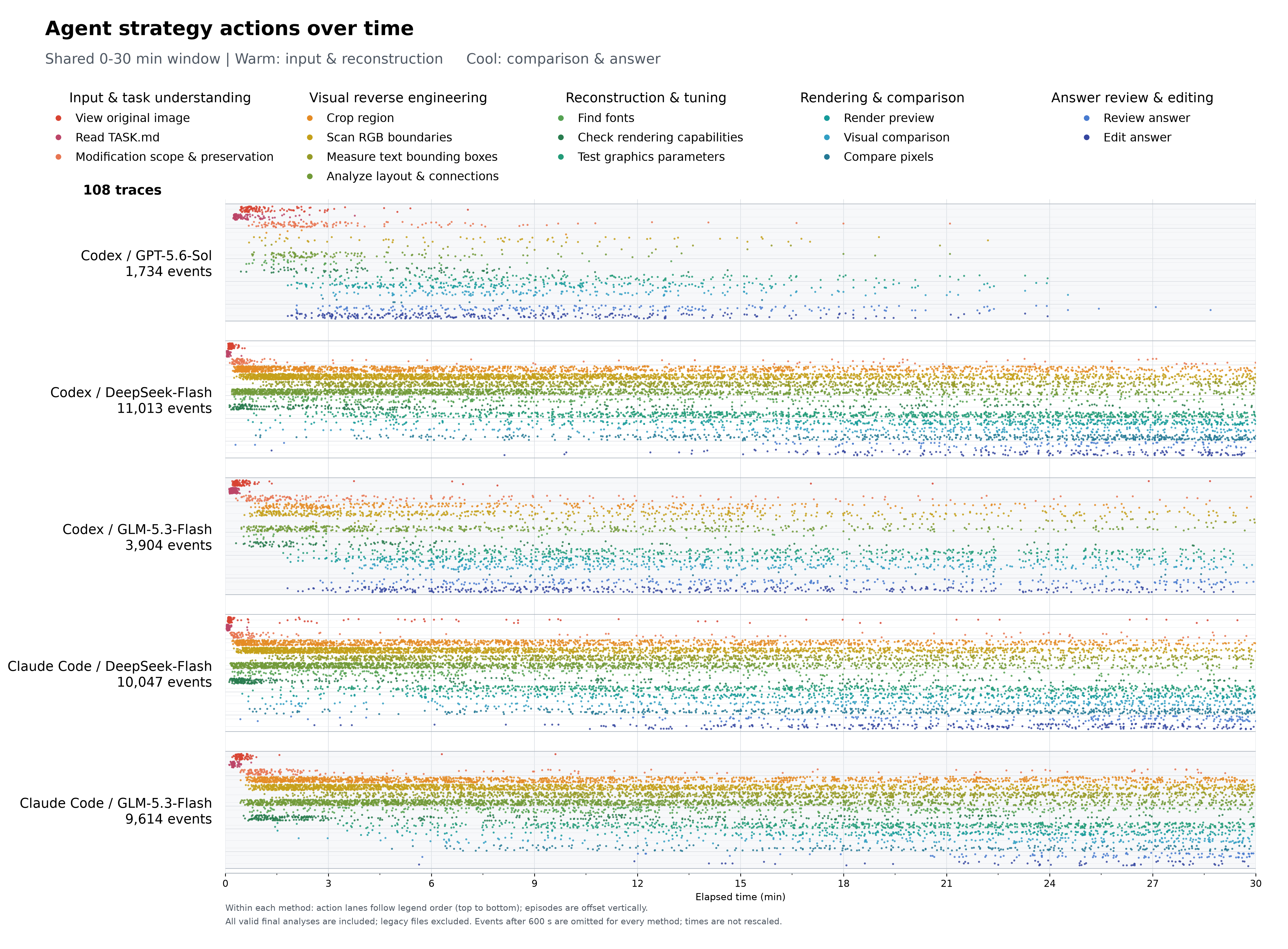}
\caption{Semantic strategy actions extracted from execution traces of five agent--model combinations on the same 108 VAO tasks. Each point corresponds to one strategy action at its observed timestamp, with colors denoting action types. The horizontal axis is normalized to the common 30-minute execution budget. Warm colors denote input understanding, reverse engineering, and reconstruction/tuning actions, whereas cool colors denote rendering, comparison, and final answer review or editing.}
\label{fig:trace-strategy}
\end{figure}

For each of the five agent--model combinations, we extract the timestamps of these strategy actions independently for each task and align them to a common 0--30 minute time axis.
Figure~\ref{fig:trace-strategy} shows the resulting traces for the same 108 tasks, allowing differences in execution behavior to be examined without conflating them with differences in task composition.
Several patterns are visible.
First, Codex--GPT-5.6-Sol shows a relatively clear progression from early task understanding and visual reverse engineering to artifact construction, rendering, comparison, and final answer review.
Its traces contain repeated rendering and verification actions distributed throughout execution, indicating an iterative workflow in which an artifact is drafted relatively early and subsequently refined against visual feedback.
This behavior is consistent with the high within-budget completion rate observed in the full experiment.

Second, Codex--DeepSeek-Flash and Codex--GLM-5.3-Flash exhibit denser and more persistent strategy activity over time.
In particular, reconstruction, tuning, rendering, and comparison actions continue throughout a large fraction of the execution window.
This suggests that the Codex harness supports iterative artifact construction across different underlying models, although the efficiency of this process varies with the model: the GLM-based configuration leaves substantially more tasks unfinished within the same time budget than the GPT-5.6-Sol configuration.

Third, the Claude Code traces show a different temporal organization.
Claude Code--DeepSeek-Flash performs substantial analysis and reverse-engineering activity before and during artifact construction, followed by repeated rendering and comparison, but the overall process remains more prolonged.
Claude Code--GLM-5.3-Flash shows the clearest concentration of analysis-oriented activity before substantial artifact construction and comparatively little productive rendering or final-answer activity within the early portion of the budget.
This execution pattern is consistent with the very small fraction of tasks completed within 30 minutes in the full evaluation.
The contrast between the two Claude Code configurations also indicates that the same harness can lead to substantially different execution behavior depending on the underlying model.

Overall, the traces suggest that effective agent execution is characterized less by the presence of individual tool actions than by how these actions are organized temporally.
In particular, early transition from task understanding and visual analysis to artifact construction, followed by repeated render--compare--edit cycles, provides a characteristic execution pattern for agents that complete a larger fraction of tasks within the available budget.
Conversely, traces dominated by prolonged analysis or reconstruction with limited time left for rendering and final revision are more likely to terminate before producing a completed answer.
The trace analysis therefore provides behavioral evidence for the role of harness--model compatibility in VAO execution: the same nominal editing capability can result in substantially different practical reliability when the interaction pattern does not efficiently support the task workflow.
Because action frequency alone does not measure task progress, we use the trace analysis jointly with the time-dependent completion curves below rather than interpreting point density as a direct performance measure.

\begin{figure}[t]
\centering
\includegraphics[width=0.8\linewidth]{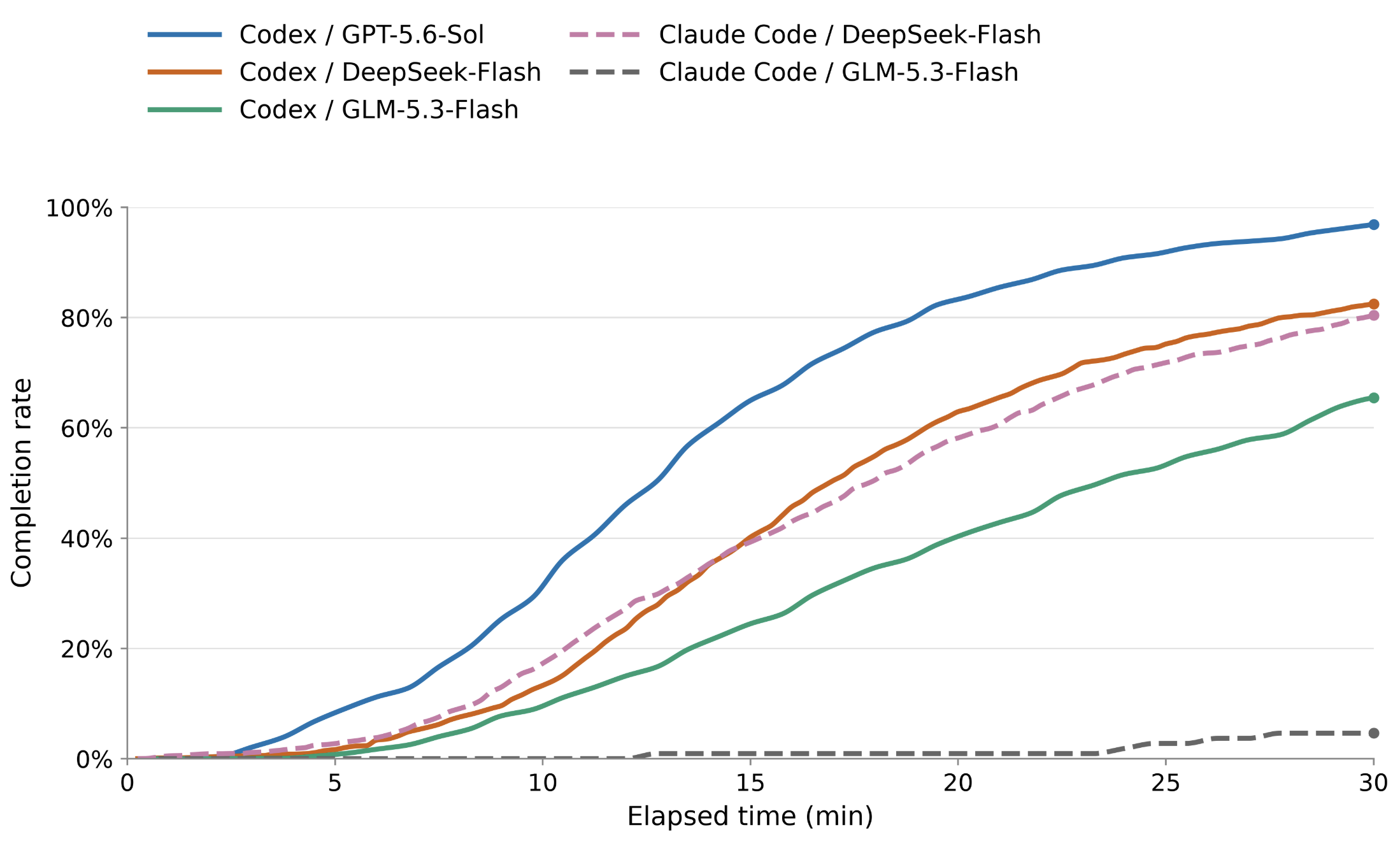}
\caption{Cumulative within-budget completion rate over elapsed execution time for five agent--model combinations on the full VAO-Bench evaluation. Each curve shows the fraction of tasks that have completed by the corresponding elapsed time; the 30-minute endpoint corresponds to the reported within-budget completion rate.}
\label{fig:completion-over-time}
\end{figure}

Figure~\ref{fig:completion-over-time} summarizes the corresponding reliability dynamics over the full evaluation set by plotting the cumulative completion rate against elapsed time.
The curves differ substantially not only in their final completion rates but also in how quickly successful executions accumulate.
Codex--GPT-5.6-Sol begins completing tasks relatively early and continues to accumulate completed tasks throughout the budget, reaching a completion rate close to the final 96.9\% reported in Table~\ref{tab:exp-main}.
Codex--DeepSeek-Flash and Claude Code--DeepSeek-Flash show slower but sustained growth, ending below 83\% within the 30-minute budget.
Codex--GLM-5.3-Flash increases more gradually and reaches 65.6\% by the deadline.
In contrast, Claude Code--GLM-5.3-Flash remains near zero for most of the execution window and only completes a small number of tasks near the end, resulting in a 4.58\% within-budget completion rate.
The shape of these curves complements the trace analysis.
A productive execution strategy should not only eventually produce a valid artifact, but should also allocate sufficient time to artifact construction, rendering, verification, and finalization before the deadline.
The large differences in the temporal accumulation of completed tasks therefore show that reliability under a fixed execution budget is strongly affected by execution dynamics, in addition to the underlying model's editing capability.

\begingroup\small
\begin{longtable}{lrrrr}
\caption{A1 reconstruction by rendering mode. SSIM and PSNR use successfully rendered outputs; SSIM-0 includes failures.}\\
\toprule
Mode & Success/total & SSIM & PSNR & SSIM-0\\\midrule
\endfirsthead
\toprule
Mode & Success/total & SSIM & PSNR & SSIM-0\\\midrule
\endhead
\bottomrule\endfoot
HTML & 106/106 & .8147 & 18.44 & .8147\\
DOT & 32/46 & .7787 & 15.39 & .5417\\
TikZ & 25/26 & .7159 & 14.33 & .6883\\
CircuitikZ & 31/33 & .8809 & 15.63 & .8275\\
Python & 97/105 & .7740 & 15.20 & .7151\\
\end{longtable}\endgroup

\section{Real Tasks and Case Studies}
\label{app:qualitative}

We present three representative VAO tasks to illustrate how instruction-guided editing can range from direct local modification to coordinated updates across dependent elements. The selected cases cover a scientific document, an interactive data panel, and a system architecture diagram, and collectively illustrate different forms of value propagation, state synchronization, and component-level editing. Each case shows the source image, reference target, and outputs from four rotating methods under the same image-only evaluation protocol.

\newcommand{\appdomaincase}[3]{%
\ifnum#3>2\clearpage\fi
\subsection{Representative Case #3}\label{app:domain-case-#3}
Task instruction: ``#2''.
\ifnum#3=1\begin{table}[H]\centering\footnotesize
\caption{Point-level scoring example for the same outputs: business web interface.}
\label{tab:app-domain-score-example}
\begin{tabular}{lrrrrr}\toprule
Method & Direct & Derived & Invariant & VQ & DQ\\\midrule
GPT-Image-2 & 1/1 & -- & 2/2 & 5 & 3\\
Seedream-5.0-Pro & 1/1 & -- & 2/2 & 5 & 5\\
GPT-5.6-Sol & 1/1 & -- & 2/2 & 5 & 5\\
DeepSeek-Flash & 1/1 & -- & 2/2 & 5 & 5\\
\bottomrule\end{tabular}\end{table}
\fi
\begingroup
\def\appcasewidth{0.48\linewidth}%
\ifnum#3=2\def\appcasewidth{0.43\linewidth}\fi
\setlength{\intextsep}{1pt}
\begin{figure}[H]\centering
\includegraphics[width=\appcasewidth]{case_new/domain_#3_row1_left.png}\hfill
\includegraphics[width=\appcasewidth]{case_new/domain_#3_row1_right.png}\par\vspace{1pt}
\includegraphics[width=\appcasewidth]{case_new/domain_#3_row2_left.png}\hfill
\includegraphics[width=\appcasewidth]{case_new/domain_#3_row2_right.png}\par\vspace{1pt}
\includegraphics[width=\appcasewidth]{case_new/domain_#3_row3_left.png}\hfill
\includegraphics[width=\appcasewidth]{case_new/domain_#3_row3_right.png}
\caption{Representative Case #3: source image and reference target (top row), followed by outputs from four rotating methods (middle and bottom rows).}
\label{fig:app-domain-#3}
\end{figure}
\endgroup
}

\appdomaincase{Case 2}{Change Method B at C3 in panel B from 58.2 to 62.0 and update the corresponding visible values.}{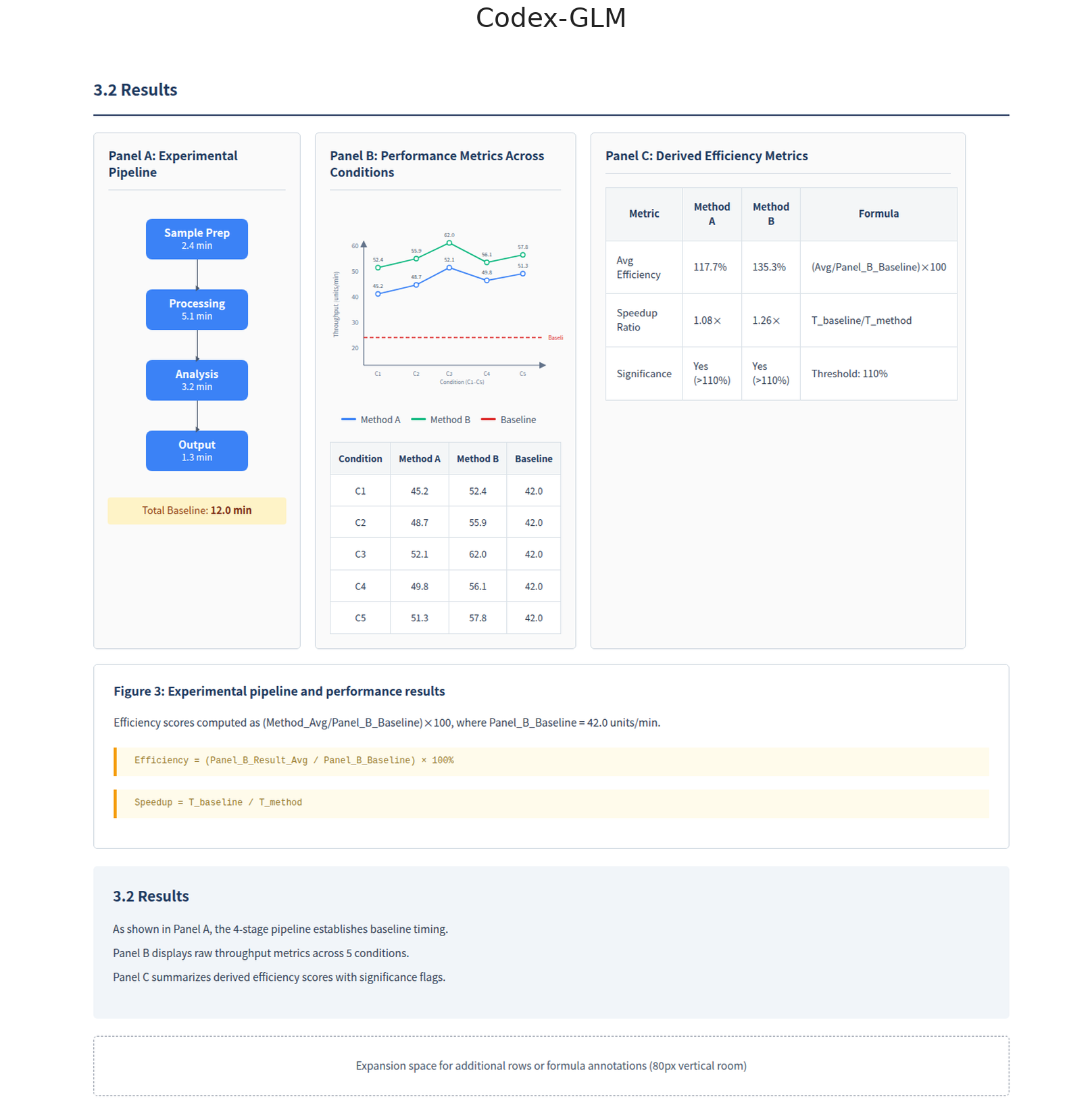}

\appdomaincase{Case 3}{Change the alert threshold from 80 to 90.}{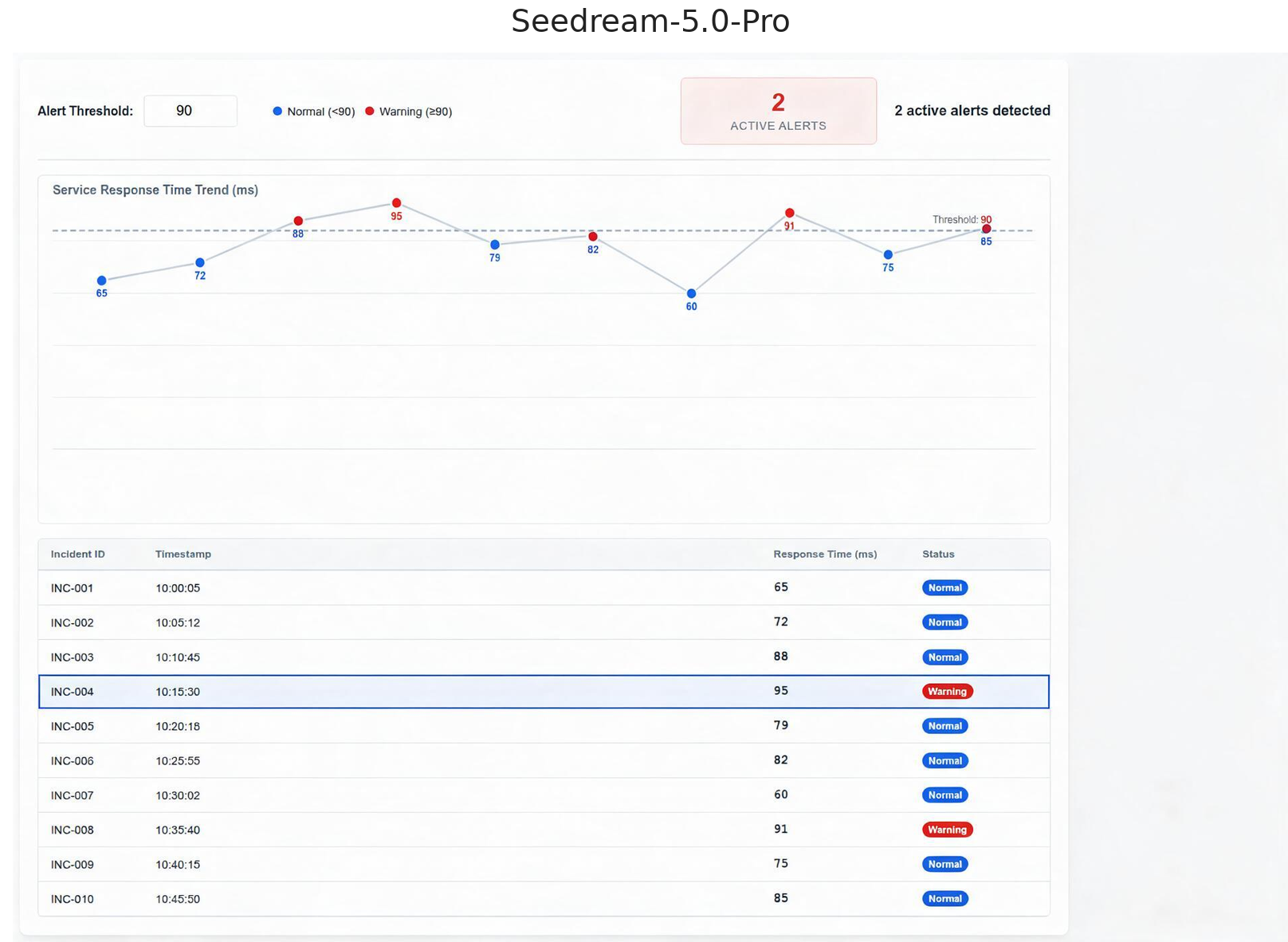}

\appdomaincase{Case 5}{Change module M1 from `Image Encoder (ViT)'' to `Visual Encoder (ViT-Base)''.}{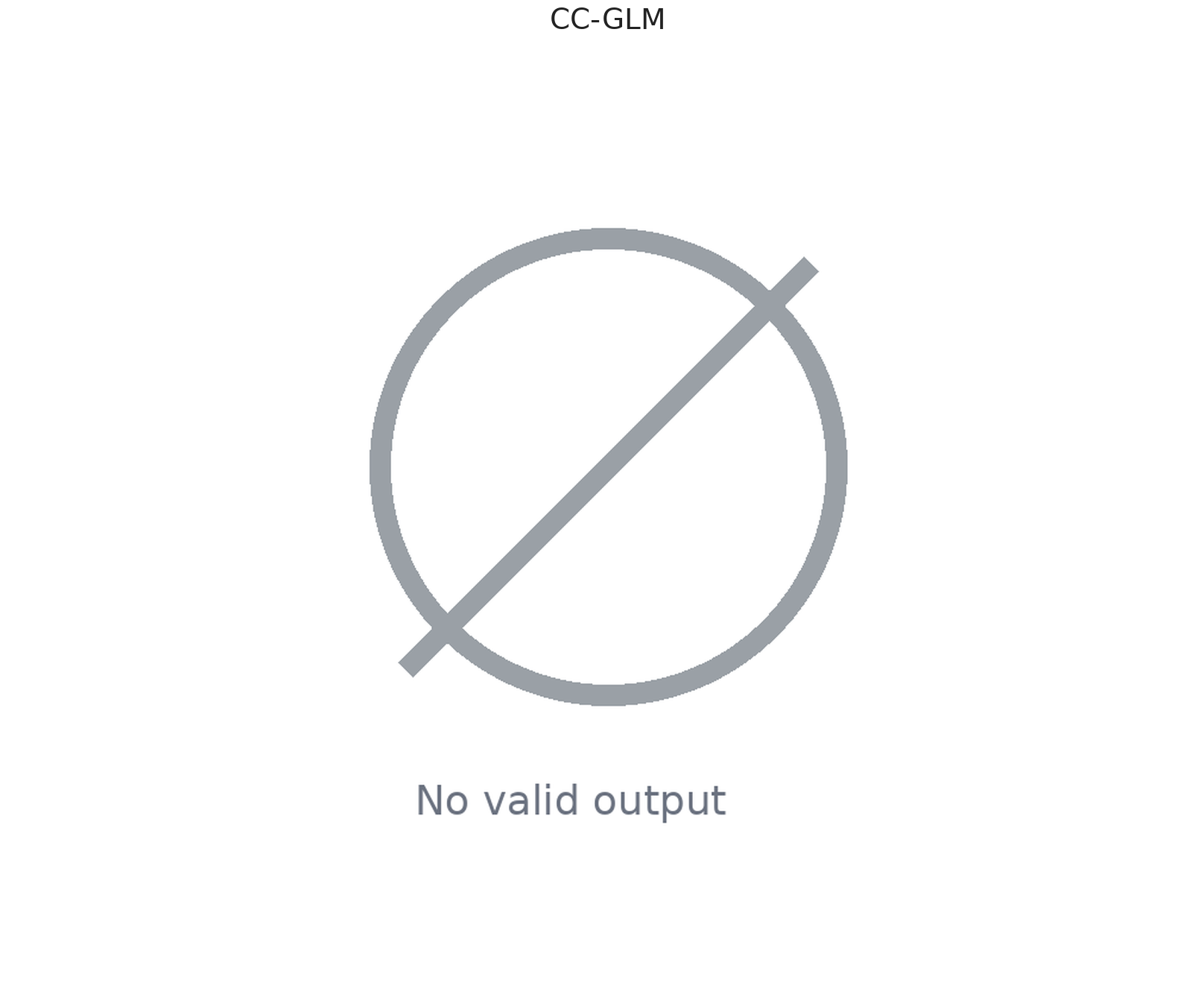}

The three cases illustrate complementary editing requirements. The scientific-document case requires a numerical change together with updates to linked visible values. The interactive-panel case requires threshold-dependent state and summary updates. The system-architecture case illustrates a component-level modification while preserving the surrounding structure and relations. Together, these examples provide concrete views of the direct, dependent, and preservation requirements evaluated by VAO-Bench. More qualitative cases and domain-specific examples are included in the Supplementary Material.

\clearpage

\section{Additional Quantitative Results}
\label{app:full-results}
\subsection{Results by Domain}
Tables~\ref{tab:app-domain-business}--\ref{tab:app-domain-chemistry} report all methods across nine domains. Each table contains auto-zero and completed-only scores to three decimal places. Method IDs follow Table~\ref{tab:app-models}. Completion rates count normal completions; both score populations include recovered outputs.

\begin{table}[!htbp]
\centering
\small
\setlength{\tabcolsep}{4pt}
\renewcommand{\arraystretch}{1.0}
\caption{Business webpages: auto-zero and completed-only results.}
\label{tab:app-domain-business}
\begin{tabular}{@{}lrr*{5}{r}@{}}
\toprule
\multicolumn{8}{c}{\textbf{Auto-zero}}\\
ID & Time (s) & Completion & Dir. & Der. & Inv. & VQ & DQ\\\midrule
M1 & 166.2 & 100.0\% & 0.689 & 0.371 & 0.951 & 4.821 & 3.740\\
M2 & 266.7 & 100.0\% & 0.810 & 0.590 & 0.944 & 4.770 & 4.010\\
M3 & 358.8 & 100.0\% & 0.785 & 0.448 & 0.971 & 4.689 & 4.005\\
M4 & 275.7 & 99.0\% & 0.795 & 0.445 & 0.932 & 4.740 & 3.974\\
M5 & 1106.4 & 83.7\% & 0.640 & 0.345 & 0.760 & 4.036 & 3.454\\
M6 & 823.8 & 100.0\% & 0.770 & 0.465 & 0.978 & 4.852 & 4.026\\
M7 & 1018.6 & 95.4\% & 0.752 & 0.445 & 0.978 & 4.816 & 4.026\\
M8 & 1165.8 & 83.2\% & 0.765 & 0.442 & 0.949 & 4.383 & 3.806\\
M9 & 876.6 & 98.5\% & 0.780 & 0.471 & 0.980 & 4.857 & 4.051\\
M10 & 1800.3 & 0.0\% & 0.240 & 0.143 & 0.371 & 2.083 & 1.583\\
\addlinespace[3pt]
\multicolumn{8}{c}{\textbf{Completed-only}}\\
ID & Time (s) & Completion & Dir. & Der. & Inv. & VQ & DQ\\\midrule
M1 & 166.2 & 100.0\% & 0.689 & 0.371 & 0.951 & 4.821 & 3.740\\
M2 & 266.7 & 100.0\% & 0.810 & 0.590 & 0.944 & 4.770 & 4.010\\
M3 & 358.8 & 100.0\% & 0.785 & 0.448 & 0.971 & 4.689 & 4.005\\
M4 & 275.7 & 99.0\% & 0.805 & 0.454 & 0.944 & 4.789 & 4.016\\
M5 & 1106.4 & 83.7\% & 0.816 & 0.455 & 0.935 & 4.823 & 4.128\\
M6 & 823.8 & 100.0\% & 0.770 & 0.465 & 0.978 & 4.852 & 4.026\\
M7 & 1018.6 & 95.4\% & 0.754 & 0.445 & 0.981 & 4.841 & 4.046\\
M8 & 1165.8 & 83.2\% & 0.765 & 0.442 & 0.949 & 4.383 & 3.806\\
M9 & 876.6 & 98.5\% & 0.780 & 0.471 & 0.980 & 4.857 & 4.051\\
M10 & 1800.3 & 0.0\% & 0.600 & 0.375 & 0.929 & 5.000 & 3.800\\
\bottomrule
\end{tabular}
\end{table}

\begin{table}[!htbp]
\centering
\small
\setlength{\tabcolsep}{4pt}
\renewcommand{\arraystretch}{1.0}
\caption{Scientific documents: auto-zero and completed-only results.}
\label{tab:app-domain-documents}
\begin{tabular}{@{}lrr*{5}{r}@{}}
\toprule
\multicolumn{8}{c}{\textbf{Auto-zero}}\\
ID & Time (s) & Completion & Dir. & Der. & Inv. & VQ & DQ\\\midrule
M1 & 177.9 & 100.0\% & 0.724 & 0.515 & 0.921 & 4.417 & 3.583\\
M2 & 271.8 & 100.0\% & 0.748 & 0.671 & 0.939 & 4.331 & 3.957\\
M3 & 334.2 & 100.0\% & 0.784 & 0.662 & 0.980 & 4.345 & 4.043\\
M4 & 364.5 & 99.3\% & 0.781 & 0.649 & 0.912 & 4.504 & 3.978\\
M5 & 1257.9 & 64.7\% & 0.515 & 0.416 & 0.606 & 2.986 & 2.640\\
M6 & 889.8 & 99.3\% & 0.811 & 0.688 & 0.986 & 4.561 & 4.043\\
M7 & 1448.2 & 61.9\% & 0.744 & 0.563 & 0.941 & 4.338 & 3.827\\
M8 & 1356.6 & 61.9\% & 0.748 & 0.584 & 0.898 & 3.849 & 3.611\\
M9 & 1414.8 & 66.2\% & 0.744 & 0.602 & 0.921 & 4.338 & 3.878\\
M10 & 1800.3 & 0.0\% & 0.037 & 0.048 & 0.167 & 0.583 & 0.583\\
\addlinespace[3pt]
\multicolumn{8}{c}{\textbf{Completed-only}}\\
ID & Time (s) & Completion & Dir. & Der. & Inv. & VQ & DQ\\\midrule
M1 & 177.9 & 100.0\% & 0.724 & 0.515 & 0.921 & 4.417 & 3.583\\
M2 & 271.8 & 100.0\% & 0.748 & 0.671 & 0.939 & 4.331 & 3.957\\
M3 & 334.2 & 100.0\% & 0.784 & 0.662 & 0.980 & 4.345 & 4.043\\
M4 & 364.5 & 99.3\% & 0.789 & 0.658 & 0.918 & 4.536 & 4.007\\
M5 & 1257.9 & 64.7\% & 0.818 & 0.686 & 0.961 & 4.611 & 4.078\\
M6 & 889.8 & 99.3\% & 0.811 & 0.688 & 0.986 & 4.561 & 4.043\\
M7 & 1448.2 & 61.9\% & 0.786 & 0.602 & 0.984 & 4.534 & 4.000\\
M8 & 1356.6 & 61.9\% & 0.776 & 0.603 & 0.934 & 4.023 & 3.774\\
M9 & 1414.8 & 66.2\% & 0.781 & 0.626 & 0.971 & 4.568 & 4.083\\
M10 & 1800.3 & 0.0\% & 0.250 & 0.333 & 1.000 & 3.500 & 3.500\\
\bottomrule
\end{tabular}
\end{table}

\begin{table}[!htbp]
\centering
\small
\setlength{\tabcolsep}{4pt}
\renewcommand{\arraystretch}{1.0}
\caption{Interactive dashboards: auto-zero and completed-only results.}
\label{tab:app-domain-dashboards}
\begin{tabular}{@{}lrr*{5}{r}@{}}
\toprule
\multicolumn{8}{c}{\textbf{Auto-zero}}\\
ID & Time (s) & Completion & Dir. & Der. & Inv. & VQ & DQ\\\midrule
M1 & 160.5 & 100.0\% & 0.702 & 0.381 & 0.956 & 4.651 & 3.413\\
M2 & 260.1 & 100.0\% & 0.808 & 0.570 & 0.958 & 4.607 & 3.817\\
M3 & 337.8 & 100.0\% & 0.740 & 0.492 & 0.986 & 4.539 & 3.869\\
M4 & 342.0 & 100.0\% & 0.743 & 0.495 & 0.929 & 4.723 & 3.885\\
M5 & 1087.2 & 77.0\% & 0.621 & 0.455 & 0.718 & 3.618 & 3.204\\
M6 & 901.5 & 97.9\% & 0.735 & 0.517 & 0.991 & 4.639 & 3.979\\
M7 & 992.5 & 95.3\% & 0.738 & 0.505 & 0.977 & 4.670 & 3.895\\
M8 & 1328.4 & 65.4\% & 0.798 & 0.595 & 0.929 & 4.225 & 3.927\\
M9 & 968.0 & 94.2\% & 0.745 & 0.505 & 0.986 & 4.702 & 3.916\\
M10 & 1800.3 & 0.0\% & 0.148 & 0.174 & 0.457 & 2.000 & 1.500\\
\addlinespace[3pt]
\multicolumn{8}{c}{\textbf{Completed-only}}\\
ID & Time (s) & Completion & Dir. & Der. & Inv. & VQ & DQ\\\midrule
M1 & 160.5 & 100.0\% & 0.702 & 0.381 & 0.956 & 4.651 & 3.413\\
M2 & 260.1 & 100.0\% & 0.808 & 0.570 & 0.958 & 4.607 & 3.817\\
M3 & 337.8 & 100.0\% & 0.740 & 0.492 & 0.986 & 4.539 & 3.869\\
M4 & 342.0 & 100.0\% & 0.743 & 0.495 & 0.929 & 4.723 & 3.885\\
M5 & 1087.2 & 77.0\% & 0.837 & 0.608 & 0.941 & 4.701 & 4.163\\
M6 & 901.5 & 97.9\% & 0.735 & 0.517 & 0.991 & 4.639 & 3.979\\
M7 & 992.5 & 95.3\% & 0.745 & 0.506 & 0.989 & 4.720 & 3.937\\
M8 & 1328.4 & 65.4\% & 0.804 & 0.595 & 0.937 & 4.270 & 3.968\\
M9 & 968.0 & 94.2\% & 0.745 & 0.505 & 0.986 & 4.702 & 3.916\\
M10 & 1800.3 & 0.0\% & 0.308 & 0.364 & 1.000 & 4.800 & 3.600\\
\bottomrule
\end{tabular}
\end{table}

\begin{table}[!htbp]
\centering
\small
\setlength{\tabcolsep}{4pt}
\renewcommand{\arraystretch}{1.0}
\caption{Flowcharts and workflows: auto-zero and completed-only results.}
\label{tab:app-domain-flowcharts}
\begin{tabular}{@{}lrr*{5}{r}@{}}
\toprule
\multicolumn{8}{c}{\textbf{Auto-zero}}\\
ID & Time (s) & Completion & Dir. & Der. & Inv. & VQ & DQ\\\midrule
M1 & 170.7 & 100.0\% & 0.721 & 0.632 & 0.898 & 4.240 & 3.969\\
M2 & 150.3 & 59.4\% & 0.478 & 0.443 & 0.580 & 2.642 & 2.576\\
M3 & 362.4 & 98.3\% & 0.803 & 0.724 & 0.891 & 3.852 & 3.974\\
M4 & 279.6 & 96.5\% & 0.771 & 0.710 & 0.815 & 3.747 & 3.664\\
M5 & 1288.2 & 44.1\% & 0.367 & 0.329 & 0.394 & 1.847 & 1.939\\
M6 & 1153.8 & 88.6\% & 0.812 & 0.774 & 0.913 & 4.096 & 4.175\\
M7 & 1366.5 & 60.3\% & 0.555 & 0.557 & 0.679 & 3.140 & 3.070\\
M8 & 1407.6 & 48.0\% & 0.666 & 0.632 & 0.748 & 3.262 & 3.432\\
M9 & 1380.2 & 64.6\% & 0.607 & 0.605 & 0.740 & 3.428 & 3.376\\
M10 & 1800.3 & 0.0\% & 0.077 & 0.091 & 0.108 & 0.667 & 0.833\\
\addlinespace[3pt]
\multicolumn{8}{c}{\textbf{Completed-only}}\\
ID & Time (s) & Completion & Dir. & Der. & Inv. & VQ & DQ\\\midrule
M1 & 170.7 & 100.0\% & 0.721 & 0.632 & 0.898 & 4.240 & 3.969\\
M2 & 150.3 & 59.4\% & 0.785 & 0.704 & 0.968 & 4.449 & 4.338\\
M3 & 362.4 & 98.3\% & 0.827 & 0.747 & 0.911 & 3.920 & 4.044\\
M4 & 279.6 & 96.5\% & 0.795 & 0.733 & 0.844 & 3.882 & 3.796\\
M5 & 1288.2 & 44.1\% & 0.832 & 0.761 & 0.910 & 4.188 & 4.396\\
M6 & 1153.8 & 88.6\% & 0.827 & 0.790 & 0.926 & 4.150 & 4.230\\
M7 & 1366.5 & 60.3\% & 0.741 & 0.719 & 0.937 & 4.331 & 4.235\\
M8 & 1407.6 & 48.0\% & 0.774 & 0.739 & 0.873 & 3.831 & 4.031\\
M9 & 1380.2 & 64.6\% & 0.743 & 0.731 & 0.932 & 4.386 & 4.318\\
M10 & 1800.3 & 0.0\% & 0.667 & 1.000 & 0.667 & 4.000 & 5.000\\
\bottomrule
\end{tabular}
\end{table}

\begin{table}[!htbp]
\centering
\small
\setlength{\tabcolsep}{4pt}
\renewcommand{\arraystretch}{1.0}
\caption{Systems and architecture diagrams: auto-zero and completed-only results.}
\label{tab:app-domain-architecture}
\begin{tabular}{@{}lrr*{5}{r}@{}}
\toprule
\multicolumn{8}{c}{\textbf{Auto-zero}}\\
ID & Time (s) & Completion & Dir. & Der. & Inv. & VQ & DQ\\\midrule
M1 & 156.0 & 100.0\% & 0.798 & 0.628 & 0.820 & 4.246 & 4.246\\
M2 & 239.7 & 100.0\% & 0.813 & 0.660 & 0.960 & 4.373 & 4.470\\
M3 & 290.4 & 88.1\% & 0.786 & 0.654 & 0.719 & 2.448 & 3.761\\
M4 & 511.2 & 85.8\% & 0.739 & 0.612 & 0.716 & 3.239 & 3.657\\
M5 & 1610.4 & 23.9\% & 0.179 & 0.165 & 0.201 & 0.881 & 1.060\\
M6 & 808.5 & 98.5\% & 0.872 & 0.734 & 0.903 & 3.806 & 4.508\\
M7 & 1244.8 & 67.2\% & 0.728 & 0.601 & 0.785 & 3.724 & 3.910\\
M8 & 1380.9 & 58.2\% & 0.802 & 0.702 & 0.794 & 3.351 & 3.955\\
M9 & 1368.6 & 61.9\% & 0.685 & 0.564 & 0.775 & 3.590 & 3.791\\
M10 & 1800.3 & 0.0\% & 0.000 & 0.000 & 0.000 & 0.000 & 0.000\\
\addlinespace[3pt]
\multicolumn{8}{c}{\textbf{Completed-only}}\\
ID & Time (s) & Completion & Dir. & Der. & Inv. & VQ & DQ\\\midrule
M1 & 156.0 & 100.0\% & 0.798 & 0.628 & 0.820 & 4.246 & 4.246\\
M2 & 239.7 & 100.0\% & 0.813 & 0.660 & 0.960 & 4.373 & 4.470\\
M3 & 290.4 & 88.1\% & 0.882 & 0.724 & 0.820 & 2.780 & 4.271\\
M4 & 511.2 & 85.8\% & 0.892 & 0.757 & 0.842 & 3.774 & 4.261\\
M5 & 1610.4 & 23.9\% & 0.821 & 0.756 & 0.852 & 3.688 & 4.438\\
M6 & 808.5 & 98.5\% & 0.875 & 0.734 & 0.908 & 3.835 & 4.541\\
M7 & 1244.8 & 67.2\% & 0.827 & 0.689 & 0.913 & 4.265 & 4.479\\
M8 & 1380.9 & 58.2\% & 0.831 & 0.725 & 0.825 & 3.508 & 4.141\\
M9 & 1368.6 & 61.9\% & 0.800 & 0.675 & 0.897 & 4.111 & 4.342\\
M10 & 1800.3 & 0.0\% & -- & -- & -- & -- & --\\
\bottomrule
\end{tabular}
\end{table}

The three overall accuracy metrics aggregate scoring points, while domain results characterize performance on different artifact types. Flowcharts combine local text, branching relations, and spatial routing; Appendix~\ref{app:qualitative} illustrates specific changes in outputs for the same task.

\begin{table}[!htbp]
\centering
\small
\setlength{\tabcolsep}{4pt}
\renewcommand{\arraystretch}{1.0}
\caption{Circuits and logic diagrams: auto-zero and completed-only results.}
\label{tab:app-domain-circuits}
\begin{tabular}{@{}lrr*{5}{r}@{}}
\toprule
\multicolumn{8}{c}{\textbf{Auto-zero}}\\
ID & Time (s) & Completion & Dir. & Der. & Inv. & VQ & DQ\\\midrule
M1 & 133.2 & 100.0\% & 0.720 & 0.659 & 0.952 & 4.385 & 3.975\\
M2 & 217.5 & 100.0\% & 0.781 & 0.731 & 0.963 & 4.435 & 4.062\\
M3 & 221.1 & 90.7\% & 0.695 & 0.659 & 0.837 & 3.106 & 3.689\\
M4 & 545.7 & 84.5\% & 0.649 & 0.560 & 0.743 & 3.410 & 3.280\\
M5 & 1644.9 & 18.0\% & 0.141 & 0.121 & 0.160 & 0.814 & 0.758\\
M6 & 602.4 & 95.0\% & 0.781 & 0.771 & 0.925 & 4.118 & 4.143\\
M7 & 950.2 & 87.6\% & 0.727 & 0.686 & 0.905 & 4.168 & 4.106\\
M8 & 1290.9 & 66.5\% & 0.704 & 0.682 & 0.873 & 3.404 & 3.584\\
M9 & 1083.3 & 80.1\% & 0.720 & 0.668 & 0.877 & 3.907 & 3.708\\
M10 & 1800.3 & 0.0\% & 0.000 & 0.000 & 0.000 & 0.000 & 0.000\\
\addlinespace[3pt]
\multicolumn{8}{c}{\textbf{Completed-only}}\\
ID & Time (s) & Completion & Dir. & Der. & Inv. & VQ & DQ\\\midrule
M1 & 133.2 & 100.0\% & 0.720 & 0.659 & 0.952 & 4.385 & 3.975\\
M2 & 217.5 & 100.0\% & 0.781 & 0.731 & 0.963 & 4.435 & 4.062\\
M3 & 221.1 & 90.7\% & 0.774 & 0.742 & 0.927 & 3.425 & 4.069\\
M4 & 545.7 & 84.5\% & 0.786 & 0.698 & 0.894 & 4.037 & 3.882\\
M5 & 1644.9 & 18.0\% & 0.815 & 0.730 & 0.961 & 4.517 & 4.207\\
M6 & 602.4 & 95.0\% & 0.805 & 0.785 & 0.959 & 4.305 & 4.331\\
M7 & 950.2 & 87.6\% & 0.774 & 0.725 & 0.958 & 4.415 & 4.349\\
M8 & 1290.9 & 66.5\% & 0.718 & 0.697 & 0.894 & 3.490 & 3.675\\
M9 & 1083.3 & 80.1\% & 0.786 & 0.720 & 0.961 & 4.308 & 4.089\\
M10 & 1800.3 & 0.0\% & -- & -- & -- & -- & --\\
\bottomrule
\end{tabular}
\end{table}

\begin{table}[!htbp]
\centering
\small
\setlength{\tabcolsep}{4pt}
\renewcommand{\arraystretch}{1.0}
\caption{Scientific statistical charts: auto-zero and completed-only results.}
\label{tab:app-domain-charts}
\begin{tabular}{@{}lrr*{5}{r}@{}}
\toprule
\multicolumn{8}{c}{\textbf{Auto-zero}}\\
ID & Time (s) & Completion & Dir. & Der. & Inv. & VQ & DQ\\\midrule
M1 & 166.8 & 100.0\% & 0.815 & 0.502 & 0.949 & 4.258 & 3.450\\
M2 & 250.8 & 100.0\% & 0.861 & 0.648 & 0.966 & 4.291 & 3.907\\
M3 & 282.3 & 99.3\% & 0.904 & 0.736 & 0.981 & 4.046 & 4.311\\
M4 & 267.6 & 99.3\% & 0.911 & 0.728 & 0.928 & 3.861 & 4.066\\
M5 & 1435.2 & 50.3\% & 0.457 & 0.347 & 0.487 & 2.086 & 2.093\\
M6 & 860.4 & 99.3\% & 0.924 & 0.736 & 0.992 & 4.404 & 4.364\\
M7 & 917.1 & 94.0\% & 0.914 & 0.766 & 0.970 & 4.517 & 4.338\\
M8 & 1341.0 & 62.3\% & 0.834 & 0.682 & 0.877 & 3.444 & 3.762\\
M9 & 990.1 & 85.4\% & 0.904 & 0.745 & 0.992 & 4.589 & 4.344\\
M10 & 1800.3 & 0.0\% & 0.200 & 0.200 & 0.231 & 1.083 & 1.083\\
\addlinespace[3pt]
\multicolumn{8}{c}{\textbf{Completed-only}}\\
ID & Time (s) & Completion & Dir. & Der. & Inv. & VQ & DQ\\\midrule
M1 & 166.8 & 100.0\% & 0.815 & 0.502 & 0.949 & 4.258 & 3.450\\
M2 & 250.8 & 100.0\% & 0.861 & 0.648 & 0.966 & 4.291 & 3.907\\
M3 & 282.3 & 99.3\% & 0.913 & 0.749 & 0.987 & 4.073 & 4.340\\
M4 & 267.6 & 99.3\% & 0.914 & 0.728 & 0.934 & 3.887 & 4.093\\
M5 & 1435.2 & 50.3\% & 0.945 & 0.709 & 0.962 & 4.145 & 4.158\\
M6 & 860.4 & 99.3\% & 0.924 & 0.736 & 0.992 & 4.404 & 4.364\\
M7 & 917.1 & 94.0\% & 0.914 & 0.766 & 0.970 & 4.517 & 4.338\\
M8 & 1341.0 & 62.3\% & 0.875 & 0.712 & 0.920 & 3.586 & 3.917\\
M9 & 990.1 & 85.4\% & 0.904 & 0.745 & 0.992 & 4.589 & 4.344\\
M10 & 1800.3 & 0.0\% & 1.000 & 1.000 & 1.000 & 4.333 & 4.333\\
\bottomrule
\end{tabular}
\end{table}

\begin{table}[!htbp]
\centering
\small
\setlength{\tabcolsep}{4pt}
\renewcommand{\arraystretch}{1.0}
\caption{Mathematics and geometry: auto-zero and completed-only results.}
\label{tab:app-domain-mathematics}
\begin{tabular}{@{}lrr*{5}{r}@{}}
\toprule
\multicolumn{8}{c}{\textbf{Auto-zero}}\\
ID & Time (s) & Completion & Dir. & Der. & Inv. & VQ & DQ\\\midrule
M1 & 138.9 & 100.0\% & 0.664 & 0.390 & 0.877 & 4.263 & 3.359\\
M2 & 218.4 & 100.0\% & 0.794 & 0.641 & 0.912 & 4.174 & 3.701\\
M3 & 277.2 & 95.8\% & 0.918 & 0.896 & 0.942 & 4.245 & 4.677\\
M4 & 343.5 & 95.2\% & 0.876 & 0.876 & 0.817 & 3.826 & 4.198\\
M5 & 1373.1 & 46.1\% & 0.394 & 0.351 & 0.410 & 1.964 & 2.084\\
M6 & 658.5 & 99.4\% & 0.949 & 0.919 & 0.976 & 4.539 & 4.796\\
M7 & 1059.6 & 89.2\% & 0.924 & 0.896 & 0.970 & 4.545 & 4.761\\
M8 & 1184.1 & 78.4\% & 0.885 & 0.788 & 0.877 & 3.826 & 4.000\\
M9 & 1164.7 & 74.9\% & 0.903 & 0.861 & 0.974 & 4.491 & 4.677\\
M10 & 1800.3 & 0.0\% & 0.043 & 0.000 & 0.077 & 0.417 & 0.417\\
\addlinespace[3pt]
\multicolumn{8}{c}{\textbf{Completed-only}}\\
ID & Time (s) & Completion & Dir. & Der. & Inv. & VQ & DQ\\\midrule
M1 & 138.9 & 100.0\% & 0.664 & 0.390 & 0.877 & 4.263 & 3.359\\
M2 & 218.4 & 100.0\% & 0.794 & 0.641 & 0.912 & 4.174 & 3.701\\
M3 & 277.2 & 95.8\% & 0.965 & 0.932 & 0.981 & 4.431 & 4.881\\
M4 & 343.5 & 95.2\% & 0.923 & 0.923 & 0.861 & 4.019 & 4.409\\
M5 & 1373.1 & 46.1\% & 0.935 & 0.892 & 0.915 & 4.260 & 4.519\\
M6 & 658.5 & 99.4\% & 0.954 & 0.922 & 0.982 & 4.566 & 4.825\\
M7 & 1059.6 & 89.2\% & 0.941 & 0.917 & 0.990 & 4.628 & 4.848\\
M8 & 1184.1 & 78.4\% & 0.901 & 0.797 & 0.895 & 3.896 & 4.073\\
M9 & 1164.7 & 74.9\% & 0.920 & 0.875 & 0.988 & 4.545 & 4.733\\
M10 & 1800.3 & 0.0\% & 1.000 & -- & 1.000 & 5.000 & 5.000\\
\bottomrule
\end{tabular}
\end{table}

\begin{table}[!htbp]
\centering
\small
\setlength{\tabcolsep}{4pt}
\renewcommand{\arraystretch}{1.0}
\caption{Chemical diagrams: auto-zero and completed-only results.}
\label{tab:app-domain-chemistry}
\begin{tabular}{@{}lrr*{5}{r}@{}}
\toprule
\multicolumn{8}{c}{\textbf{Auto-zero}}\\
ID & Time (s) & Completion & Dir. & Der. & Inv. & VQ & DQ\\\midrule
M1 & 120.3 & 100.0\% & 0.770 & 0.706 & 0.903 & 4.785 & 4.278\\
M2 & 199.2 & 99.0\% & 0.809 & 0.752 & 0.927 & 4.566 & 4.258\\
M3 & 216.3 & 99.5\% & 0.850 & 0.752 & 0.954 & 4.688 & 4.488\\
M4 & 343.5 & 90.2\% & 0.745 & 0.636 & 0.650 & 3.668 & 3.615\\
M5 & 1490.1 & 37.6\% & 0.306 & 0.280 & 0.292 & 1.532 & 1.566\\
M6 & 528.3 & 97.6\% & 0.821 & 0.755 & 0.951 & 4.707 & 4.561\\
M7 & 801.0 & 89.8\% & 0.827 & 0.734 & 0.941 & 4.644 & 4.517\\
M8 & 1316.4 & 67.3\% & 0.765 & 0.668 & 0.784 & 3.971 & 3.795\\
M9 & 793.9 & 90.7\% & 0.842 & 0.745 & 0.966 & 4.737 & 4.473\\
M10 & 1621.2 & 41.7\% & 0.318 & 0.143 & 0.457 & 2.417 & 2.167\\
\addlinespace[3pt]
\multicolumn{8}{c}{\textbf{Completed-only}}\\
ID & Time (s) & Completion & Dir. & Der. & Inv. & VQ & DQ\\\midrule
M1 & 120.3 & 100.0\% & 0.770 & 0.706 & 0.903 & 4.785 & 4.278\\
M2 & 199.2 & 99.0\% & 0.826 & 0.773 & 0.939 & 4.611 & 4.301\\
M3 & 216.3 & 99.5\% & 0.854 & 0.752 & 0.958 & 4.711 & 4.510\\
M4 & 343.5 & 90.2\% & 0.839 & 0.711 & 0.722 & 4.065 & 4.005\\
M5 & 1490.1 & 37.6\% & 0.845 & 0.748 & 0.784 & 4.078 & 4.169\\
M6 & 528.3 & 97.6\% & 0.839 & 0.769 & 0.975 & 4.825 & 4.675\\
M7 & 801.0 & 89.8\% & 0.839 & 0.739 & 0.964 & 4.760 & 4.630\\
M8 & 1316.4 & 67.3\% & 0.833 & 0.707 & 0.855 & 4.376 & 4.183\\
M9 & 793.9 & 90.7\% & 0.851 & 0.753 & 0.981 & 4.807 & 4.540\\
M10 & 1621.2 & 41.7\% & 0.875 & 0.500 & 1.000 & 4.833 & 4.333\\
\bottomrule
\end{tabular}
\end{table}

\begin{table}[!htbp]
\centering
\small
\setlength{\tabcolsep}{6pt}
\caption{Overall scoring-point denominators. Auto-zero includes execution failures; completed-only includes successfully scored tasks. Both include successfully scored outputs recovered after timeout.}
\label{tab:app-denominators}
\begin{tabular}{llrrr}
\toprule
Method & Population & Direct & Derived & Invariant\\\midrule
GPT-Image-2 & Auto-zero & 3154 & 2416 & 4775\\
GPT-Image-2 & Completed-only & 3154 & 2416 & 4775\\
Seedream-5.0-Pro & Auto-zero & 3154 & 2416 & 4775\\
Seedream-5.0-Pro & Completed-only & 2967 & 2275 & 4477\\
GPT-5.6-Sol & Auto-zero & 3154 & 2416 & 4775\\
GPT-5.6-Sol & Completed-only & 3060 & 2348 & 4643\\
DeepSeek-Flash & Auto-zero & 3154 & 2416 & 4775\\
DeepSeek-Flash & Completed-only & 2972 & 2273 & 4514\\
GLM-5.3-Flash & Auto-zero & 3154 & 2416 & 4775\\
GLM-5.3-Flash & Completed-only & 1542 & 1174 & 2377\\
Codex-GPT & Auto-zero & 3154 & 2416 & 4775\\
Codex-GPT & Completed-only & 3126 & 2399 & 4728\\
Codex-DeepSeek & Auto-zero & 3154 & 2416 & 4775\\
Codex-DeepSeek & Completed-only & 2956 & 2275 & 4446\\
Codex-GLM & Auto-zero & 3154 & 2416 & 4775\\
Codex-GLM & Completed-only & 3009 & 2317 & 4539\\
CC-DeepSeek & Auto-zero & 3154 & 2416 & 4775\\
CC-DeepSeek & Completed-only & 2983 & 2291 & 4495\\
CC-GLM & Auto-zero & 3154 & 2416 & 4775\\
CC-GLM & Completed-only & 641 & 466 & 1014\\
\bottomrule
\end{tabular}
\end{table}

VAO-Bench contains 3,154 Direct, 2,416 Derived, and 4,775 Invariant points. Auto-zero denominators cover tasks included in the aggregation; completed-only denominators cover each method's successfully scored tasks.

\FloatBarrier

\subsection{Results by Difficulty}\label{app:difficulty-results}
Tables~\ref{tab:app-difficulty-easy}--\ref{tab:app-difficulty-hard} report results for the three difficulty levels.
\begin{table}[!htbp]
\centering
\small
\setlength{\tabcolsep}{4pt}
\renewcommand{\arraystretch}{1.0}
\caption{Easy tasks: auto-zero and completed-only results.}
\label{tab:app-difficulty-easy}
\begin{tabular}{@{}lrr*{5}{r}@{}}
\toprule
\multicolumn{8}{c}{\textbf{Auto-zero}}\\
ID & Time (s) & Completion & Dir. & Der. & Inv. & VQ & DQ\\\midrule
M1 & 150.3 & 100.0\% & 0.897 & 0.556 & 0.929 & 4.570 & 4.520\\
M2 & 210.0 & 93.3\% & 0.884 & 0.667 & 0.900 & 4.300 & 4.338\\
M3 & 307.2 & 97.0\% & 0.916 & 0.556 & 0.920 & 4.115 & 4.368\\
M4 & 339.3 & 95.3\% & 0.909 & 0.556 & 0.810 & 4.027 & 4.097\\
M5 & 1304.7 & 52.0\% & 0.501 & 0.556 & 0.457 & 2.327 & 2.363\\
M6 & 848.1 & 95.3\% & 0.933 & 0.667 & 0.946 & 4.480 & 4.592\\
M7 & 1125.4 & 79.2\% & 0.850 & 0.667 & 0.878 & 4.293 & 4.308\\
M8 & 1390.2 & 56.8\% & 0.863 & 0.667 & 0.811 & 3.752 & 3.997\\
M9 & 1161.3 & 76.5\% & 0.877 & 0.667 & 0.900 & 4.345 & 4.372\\
M10 & 1760.7 & 5.6\% & 0.209 & 0.000 & 0.286 & 1.361 & 1.222\\
\addlinespace[3pt]
\multicolumn{8}{c}{\textbf{Completed-only}}\\
ID & Time (s) & Completion & Dir. & Der. & Inv. & VQ & DQ\\\midrule
M1 & 150.3 & 100.0\% & 0.897 & 0.556 & 0.929 & 4.570 & 4.520\\
M2 & 210.0 & 93.3\% & 0.939 & 0.667 & 0.967 & 4.607 & 4.648\\
M3 & 307.2 & 97.0\% & 0.943 & 0.556 & 0.946 & 4.242 & 4.503\\
M4 & 339.3 & 95.3\% & 0.954 & 0.625 & 0.852 & 4.224 & 4.297\\
M5 & 1304.7 & 52.0\% & 0.958 & 1.000 & 0.889 & 4.474 & 4.545\\
M6 & 848.1 & 95.3\% & 0.944 & 0.667 & 0.957 & 4.540 & 4.654\\
M7 & 1125.4 & 79.2\% & 0.917 & 0.667 & 0.962 & 4.684 & 4.700\\
M8 & 1390.2 & 56.8\% & 0.920 & 0.667 & 0.867 & 4.005 & 4.267\\
M9 & 1161.3 & 76.5\% & 0.931 & 0.667 & 0.963 & 4.639 & 4.667\\
M10 & 1760.7 & 5.6\% & 0.818 & 0.000 & 1.000 & 4.900 & 4.400\\
\bottomrule
\end{tabular}
\end{table}

\begin{table}[!htbp]
\centering
\small
\setlength{\tabcolsep}{4pt}
\renewcommand{\arraystretch}{1.0}
\caption{Medium tasks: auto-zero and completed-only results.}
\label{tab:app-difficulty-medium}
\begin{tabular}{@{}lrr*{5}{r}@{}}
\toprule
\multicolumn{8}{c}{\textbf{Auto-zero}}\\
ID & Time (s) & Completion & Dir. & Der. & Inv. & VQ & DQ\\\midrule
M1 & 153.3 & 100.0\% & 0.772 & 0.546 & 0.925 & 4.413 & 3.616\\
M2 & 224.7 & 94.7\% & 0.777 & 0.640 & 0.898 & 4.176 & 3.764\\
M3 & 296.7 & 97.7\% & 0.846 & 0.656 & 0.933 & 4.036 & 4.002\\
M4 & 358.8 & 94.9\% & 0.826 & 0.619 & 0.833 & 3.993 & 3.707\\
M5 & 1348.2 & 52.2\% & 0.457 & 0.336 & 0.487 & 2.330 & 2.161\\
M6 & 801.0 & 97.7\% & 0.863 & 0.680 & 0.959 & 4.402 & 4.086\\
M7 & 1069.6 & 84.2\% & 0.818 & 0.605 & 0.916 & 4.285 & 3.989\\
M8 & 1311.0 & 67.5\% & 0.789 & 0.603 & 0.870 & 3.728 & 3.623\\
M9 & 1099.7 & 82.0\% & 0.804 & 0.600 & 0.905 & 4.255 & 3.865\\
M10 & 1780.2 & 8.3\% & 0.174 & 0.200 & 0.224 & 1.028 & 1.056\\
\addlinespace[3pt]
\multicolumn{8}{c}{\textbf{Completed-only}}\\
ID & Time (s) & Completion & Dir. & Der. & Inv. & VQ & DQ\\\midrule
M1 & 153.3 & 100.0\% & 0.772 & 0.546 & 0.925 & 4.413 & 3.616\\
M2 & 224.7 & 94.7\% & 0.820 & 0.677 & 0.950 & 4.408 & 3.972\\
M3 & 296.7 & 97.7\% & 0.866 & 0.670 & 0.954 & 4.129 & 4.094\\
M4 & 358.8 & 94.9\% & 0.870 & 0.651 & 0.878 & 4.205 & 3.905\\
M5 & 1348.2 & 52.2\% & 0.875 & 0.637 & 0.928 & 4.468 & 4.144\\
M6 & 801.0 & 97.7\% & 0.875 & 0.688 & 0.971 & 4.460 & 4.141\\
M7 & 1069.6 & 84.2\% & 0.867 & 0.641 & 0.974 & 4.541 & 4.227\\
M8 & 1311.0 & 67.5\% & 0.832 & 0.637 & 0.914 & 3.927 & 3.816\\
M9 & 1099.7 & 82.0\% & 0.859 & 0.641 & 0.971 & 4.563 & 4.145\\
M10 & 1780.2 & 8.3\% & 0.706 & 0.733 & 0.889 & 4.111 & 4.222\\
\bottomrule
\end{tabular}
\end{table}

\begin{table}[!htbp]
\centering
\small
\setlength{\tabcolsep}{4pt}
\renewcommand{\arraystretch}{1.0}
\caption{Hard tasks: auto-zero and completed-only results.}
\label{tab:app-difficulty-hard}
\begin{tabular}{@{}lrr*{5}{r}@{}}
\toprule
\multicolumn{8}{c}{\textbf{Auto-zero}}\\
ID & Time (s) & Completion & Dir. & Der. & Inv. & VQ & DQ\\\midrule
M1 & 160.2 & 100.0\% & 0.622 & 0.515 & 0.895 & 4.396 & 3.000\\
M2 & 251.7 & 93.9\% & 0.682 & 0.611 & 0.872 & 4.084 & 3.136\\
M3 & 294.6 & 96.8\% & 0.728 & 0.668 & 0.921 & 4.002 & 3.845\\
M4 & 372.0 & 93.4\% & 0.683 & 0.639 & 0.835 & 3.927 & 3.530\\
M5 & 1420.5 & 46.1\% & 0.345 & 0.313 & 0.429 & 2.018 & 1.770\\
M6 & 771.3 & 98.2\% & 0.747 & 0.709 & 0.966 & 4.400 & 4.091\\
M7 & 1039.8 & 85.2\% & 0.676 & 0.643 & 0.907 & 4.229 & 3.677\\
M8 & 1188.9 & 75.2\% & 0.713 & 0.652 & 0.883 & 3.818 & 3.582\\
M9 & 1025.1 & 83.9\% & 0.687 & 0.655 & 0.928 & 4.289 & 3.704\\
M10 & 1800.3 & 0.0\% & 0.042 & 0.040 & 0.129 & 0.694 & 0.444\\
\addlinespace[3pt]
\multicolumn{8}{c}{\textbf{Completed-only}}\\
ID & Time (s) & Completion & Dir. & Der. & Inv. & VQ & DQ\\\midrule
M1 & 160.2 & 100.0\% & 0.622 & 0.515 & 0.895 & 4.396 & 3.000\\
M2 & 251.7 & 93.9\% & 0.729 & 0.651 & 0.931 & 4.351 & 3.341\\
M3 & 294.6 & 96.8\% & 0.754 & 0.691 & 0.952 & 4.134 & 3.972\\
M4 & 372.0 & 93.4\% & 0.733 & 0.683 & 0.891 & 4.204 & 3.779\\
M5 & 1420.5 & 46.1\% & 0.767 & 0.678 & 0.941 & 4.374 & 3.837\\
M6 & 771.3 & 98.2\% & 0.751 & 0.712 & 0.971 & 4.420 & 4.110\\
M7 & 1039.8 & 85.2\% & 0.721 & 0.684 & 0.966 & 4.495 & 3.908\\
M8 & 1188.9 & 75.2\% & 0.738 & 0.675 & 0.915 & 3.953 & 3.708\\
M9 & 1025.1 & 83.9\% & 0.720 & 0.685 & 0.971 & 4.493 & 3.881\\
M10 & 1800.3 & 0.0\% & 0.312 & 0.312 & 1.000 & 5.000 & 3.200\\
\bottomrule
\end{tabular}
\end{table}

\FloatBarrier
\subsection{Results by Rendering Mode}\label{app:rendering-results}
Tables~\ref{tab:app-render-html}--\ref{tab:app-render-python} report results for the five rendering modes.
\begin{table}[!htbp]
\centering
\small
\setlength{\tabcolsep}{4pt}
\renewcommand{\arraystretch}{1.0}
\caption{HTML rendering: auto-zero and completed-only results.}
\label{tab:app-render-html}
\begin{tabular}{@{}lrr*{5}{r}@{}}
\toprule
\multicolumn{8}{c}{\textbf{Auto-zero}}\\
ID & Time (s) & Completion & Dir. & Der. & Inv. & VQ & DQ\\\midrule
M1 & 167.4 & 100.0\% & 0.703 & 0.414 & 0.944 & 4.653 & 3.580\\
M2 & 265.8 & 100.0\% & 0.793 & 0.604 & 0.948 & 4.595 & 3.926\\
M3 & 344.7 & 100.0\% & 0.768 & 0.522 & 0.979 & 4.544 & 3.966\\
M4 & 323.1 & 99.4\% & 0.772 & 0.519 & 0.925 & 4.671 & 3.943\\
M5 & 1139.4 & 76.2\% & 0.600 & 0.405 & 0.703 & 3.607 & 3.148\\
M6 & 869.4 & 99.0\% & 0.768 & 0.544 & 0.985 & 4.698 & 4.013\\
M7 & 1122.7 & 86.5\% & 0.745 & 0.499 & 0.968 & 4.637 & 3.926\\
M8 & 1275.3 & 71.1\% & 0.773 & 0.537 & 0.928 & 4.184 & 3.799\\
M9 & 1052.0 & 88.4\% & 0.757 & 0.519 & 0.966 & 4.663 & 3.956\\
M10 & 1800.3 & 0.0\% & 0.139 & 0.123 & 0.330 & 1.556 & 1.222\\
\addlinespace[3pt]
\multicolumn{8}{c}{\textbf{Completed-only}}\\
ID & Time (s) & Completion & Dir. & Der. & Inv. & VQ & DQ\\\midrule
M1 & 167.4 & 100.0\% & 0.703 & 0.414 & 0.944 & 4.653 & 3.580\\
M2 & 265.8 & 100.0\% & 0.793 & 0.604 & 0.948 & 4.595 & 3.926\\
M3 & 344.7 & 100.0\% & 0.768 & 0.522 & 0.979 & 4.544 & 3.966\\
M4 & 323.1 & 99.4\% & 0.777 & 0.524 & 0.931 & 4.698 & 3.966\\
M5 & 1139.4 & 76.2\% & 0.824 & 0.568 & 0.943 & 4.731 & 4.130\\
M6 & 869.4 & 99.0\% & 0.768 & 0.544 & 0.985 & 4.698 & 4.013\\
M7 & 1122.7 & 86.5\% & 0.759 & 0.508 & 0.985 & 4.718 & 3.994\\
M8 & 1275.3 & 71.1\% & 0.783 & 0.541 & 0.941 & 4.249 & 3.857\\
M9 & 1052.0 & 88.4\% & 0.767 & 0.524 & 0.980 & 4.726 & 4.010\\
M10 & 1800.3 & 0.0\% & 0.407 & 0.364 & 0.972 & 4.667 & 3.667\\
\bottomrule
\end{tabular}
\end{table}

\begin{table}[!htbp]
\centering
\small
\setlength{\tabcolsep}{4pt}
\renewcommand{\arraystretch}{1.0}
\caption{DOT rendering: auto-zero and completed-only results.}
\label{tab:app-render-dot}
\begin{tabular}{@{}lrr*{5}{r}@{}}
\toprule
\multicolumn{8}{c}{\textbf{Auto-zero}}\\
ID & Time (s) & Completion & Dir. & Der. & Inv. & VQ & DQ\\\midrule
M1 & 170.7 & 100.0\% & 0.721 & 0.632 & 0.898 & 4.240 & 3.969\\
M2 & 150.3 & 59.4\% & 0.478 & 0.443 & 0.580 & 2.642 & 2.576\\
M3 & 362.4 & 98.3\% & 0.803 & 0.724 & 0.891 & 3.852 & 3.974\\
M4 & 279.6 & 96.5\% & 0.771 & 0.710 & 0.815 & 3.747 & 3.664\\
M5 & 1288.2 & 44.1\% & 0.367 & 0.329 & 0.394 & 1.847 & 1.939\\
M6 & 1153.8 & 88.6\% & 0.812 & 0.774 & 0.913 & 4.096 & 4.175\\
M7 & 1366.5 & 60.3\% & 0.555 & 0.557 & 0.679 & 3.140 & 3.070\\
M8 & 1407.6 & 48.0\% & 0.666 & 0.632 & 0.748 & 3.262 & 3.432\\
M9 & 1380.2 & 64.6\% & 0.607 & 0.605 & 0.740 & 3.428 & 3.376\\
M10 & 1800.3 & 0.0\% & 0.077 & 0.091 & 0.108 & 0.667 & 0.833\\
\addlinespace[3pt]
\multicolumn{8}{c}{\textbf{Completed-only}}\\
ID & Time (s) & Completion & Dir. & Der. & Inv. & VQ & DQ\\\midrule
M1 & 170.7 & 100.0\% & 0.721 & 0.632 & 0.898 & 4.240 & 3.969\\
M2 & 150.3 & 59.4\% & 0.785 & 0.704 & 0.968 & 4.449 & 4.338\\
M3 & 362.4 & 98.3\% & 0.827 & 0.747 & 0.911 & 3.920 & 4.044\\
M4 & 279.6 & 96.5\% & 0.795 & 0.733 & 0.844 & 3.882 & 3.796\\
M5 & 1288.2 & 44.1\% & 0.832 & 0.761 & 0.910 & 4.188 & 4.396\\
M6 & 1153.8 & 88.6\% & 0.827 & 0.790 & 0.926 & 4.150 & 4.230\\
M7 & 1366.5 & 60.3\% & 0.741 & 0.719 & 0.937 & 4.331 & 4.235\\
M8 & 1407.6 & 48.0\% & 0.774 & 0.739 & 0.873 & 3.831 & 4.031\\
M9 & 1380.2 & 64.6\% & 0.743 & 0.731 & 0.932 & 4.386 & 4.318\\
M10 & 1800.3 & 0.0\% & 0.667 & 1.000 & 0.667 & 4.000 & 5.000\\
\bottomrule
\end{tabular}
\end{table}

\begin{table}[!htbp]
\centering
\small
\setlength{\tabcolsep}{4pt}
\renewcommand{\arraystretch}{1.0}
\caption{TikZ rendering: auto-zero and completed-only results.}
\label{tab:app-render-tikz}
\begin{tabular}{@{}lrr*{5}{r}@{}}
\toprule
\multicolumn{8}{c}{\textbf{Auto-zero}}\\
ID & Time (s) & Completion & Dir. & Der. & Inv. & VQ & DQ\\\midrule
M1 & 156.0 & 100.0\% & 0.798 & 0.628 & 0.820 & 4.246 & 4.246\\
M2 & 239.7 & 100.0\% & 0.813 & 0.660 & 0.960 & 4.373 & 4.470\\
M3 & 290.4 & 88.1\% & 0.786 & 0.654 & 0.719 & 2.448 & 3.761\\
M4 & 511.2 & 85.8\% & 0.739 & 0.612 & 0.716 & 3.239 & 3.657\\
M5 & 1610.4 & 23.9\% & 0.179 & 0.165 & 0.201 & 0.881 & 1.060\\
M6 & 808.5 & 98.5\% & 0.872 & 0.734 & 0.903 & 3.806 & 4.508\\
M7 & 1244.8 & 67.2\% & 0.728 & 0.601 & 0.785 & 3.724 & 3.910\\
M8 & 1380.9 & 58.2\% & 0.802 & 0.702 & 0.794 & 3.351 & 3.955\\
M9 & 1368.6 & 61.9\% & 0.685 & 0.564 & 0.775 & 3.590 & 3.791\\
M10 & 1800.3 & 0.0\% & 0.000 & 0.000 & 0.000 & 0.000 & 0.000\\
\addlinespace[3pt]
\multicolumn{8}{c}{\textbf{Completed-only}}\\
ID & Time (s) & Completion & Dir. & Der. & Inv. & VQ & DQ\\\midrule
M1 & 156.0 & 100.0\% & 0.798 & 0.628 & 0.820 & 4.246 & 4.246\\
M2 & 239.7 & 100.0\% & 0.813 & 0.660 & 0.960 & 4.373 & 4.470\\
M3 & 290.4 & 88.1\% & 0.882 & 0.724 & 0.820 & 2.780 & 4.271\\
M4 & 511.2 & 85.8\% & 0.892 & 0.757 & 0.842 & 3.774 & 4.261\\
M5 & 1610.4 & 23.9\% & 0.821 & 0.756 & 0.852 & 3.688 & 4.438\\
M6 & 808.5 & 98.5\% & 0.875 & 0.734 & 0.908 & 3.835 & 4.541\\
M7 & 1244.8 & 67.2\% & 0.827 & 0.689 & 0.913 & 4.265 & 4.479\\
M8 & 1380.9 & 58.2\% & 0.831 & 0.725 & 0.825 & 3.508 & 4.141\\
M9 & 1368.6 & 61.9\% & 0.800 & 0.675 & 0.897 & 4.111 & 4.342\\
M10 & 1800.3 & 0.0\% & -- & -- & -- & -- & --\\
\bottomrule
\end{tabular}
\end{table}

\begin{table}[!htbp]
\centering
\small
\setlength{\tabcolsep}{4pt}
\renewcommand{\arraystretch}{1.0}
\caption{CircuitikZ rendering: auto-zero and completed-only results.}
\label{tab:app-render-circuitikz}
\begin{tabular}{@{}lrr*{5}{r}@{}}
\toprule
\multicolumn{8}{c}{\textbf{Auto-zero}}\\
ID & Time (s) & Completion & Dir. & Der. & Inv. & VQ & DQ\\\midrule
M1 & 133.2 & 100.0\% & 0.720 & 0.659 & 0.952 & 4.385 & 3.975\\
M2 & 217.5 & 100.0\% & 0.781 & 0.731 & 0.963 & 4.435 & 4.062\\
M3 & 221.1 & 90.7\% & 0.695 & 0.659 & 0.837 & 3.106 & 3.689\\
M4 & 545.7 & 84.5\% & 0.649 & 0.560 & 0.743 & 3.410 & 3.280\\
M5 & 1644.9 & 18.0\% & 0.141 & 0.121 & 0.160 & 0.814 & 0.758\\
M6 & 602.4 & 95.0\% & 0.781 & 0.771 & 0.925 & 4.118 & 4.143\\
M7 & 950.2 & 87.6\% & 0.727 & 0.686 & 0.905 & 4.168 & 4.106\\
M8 & 1290.9 & 66.5\% & 0.704 & 0.682 & 0.873 & 3.404 & 3.584\\
M9 & 1083.3 & 80.1\% & 0.720 & 0.668 & 0.877 & 3.907 & 3.708\\
M10 & 1800.3 & 0.0\% & 0.000 & 0.000 & 0.000 & 0.000 & 0.000\\
\addlinespace[3pt]
\multicolumn{8}{c}{\textbf{Completed-only}}\\
ID & Time (s) & Completion & Dir. & Der. & Inv. & VQ & DQ\\\midrule
M1 & 133.2 & 100.0\% & 0.720 & 0.659 & 0.952 & 4.385 & 3.975\\
M2 & 217.5 & 100.0\% & 0.781 & 0.731 & 0.963 & 4.435 & 4.062\\
M3 & 221.1 & 90.7\% & 0.774 & 0.742 & 0.927 & 3.425 & 4.069\\
M4 & 545.7 & 84.5\% & 0.786 & 0.698 & 0.894 & 4.037 & 3.882\\
M5 & 1644.9 & 18.0\% & 0.815 & 0.730 & 0.961 & 4.517 & 4.207\\
M6 & 602.4 & 95.0\% & 0.805 & 0.785 & 0.959 & 4.305 & 4.331\\
M7 & 950.2 & 87.6\% & 0.774 & 0.725 & 0.958 & 4.415 & 4.349\\
M8 & 1290.9 & 66.5\% & 0.718 & 0.697 & 0.894 & 3.490 & 3.675\\
M9 & 1083.3 & 80.1\% & 0.786 & 0.720 & 0.961 & 4.308 & 4.089\\
M10 & 1800.3 & 0.0\% & -- & -- & -- & -- & --\\
\bottomrule
\end{tabular}
\end{table}

\begin{table}[!htbp]
\centering
\small
\setlength{\tabcolsep}{4pt}
\renewcommand{\arraystretch}{1.0}
\caption{Python rendering: auto-zero and completed-only results.}
\label{tab:app-render-python}
\begin{tabular}{@{}lrr*{5}{r}@{}}
\toprule
\multicolumn{8}{c}{\textbf{Auto-zero}}\\
ID & Time (s) & Completion & Dir. & Der. & Inv. & VQ & DQ\\\midrule
M1 & 139.8 & 100.0\% & 0.749 & 0.539 & 0.908 & 4.466 & 3.746\\
M2 & 220.2 & 99.6\% & 0.819 & 0.684 & 0.934 & 4.361 & 3.979\\
M3 & 254.7 & 98.3\% & 0.888 & 0.795 & 0.958 & 4.361 & 4.497\\
M4 & 321.6 & 94.5\% & 0.836 & 0.744 & 0.782 & 3.774 & 3.931\\
M5 & 1437.0 & 44.0\% & 0.379 & 0.324 & 0.385 & 1.830 & 1.883\\
M6 & 665.7 & 98.7\% & 0.893 & 0.804 & 0.970 & 4.566 & 4.579\\
M7 & 917.1 & 90.8\% & 0.884 & 0.797 & 0.959 & 4.575 & 4.543\\
M8 & 1281.3 & 69.4\% & 0.824 & 0.712 & 0.840 & 3.772 & 3.851\\
M9 & 968.9 & 84.1\% & 0.880 & 0.783 & 0.976 & 4.616 & 4.501\\
M10 & 1740.6 & 13.9\% & 0.186 & 0.109 & 0.248 & 1.306 & 1.222\\
\addlinespace[3pt]
\multicolumn{8}{c}{\textbf{Completed-only}}\\
ID & Time (s) & Completion & Dir. & Der. & Inv. & VQ & DQ\\\midrule
M1 & 139.8 & 100.0\% & 0.749 & 0.539 & 0.908 & 4.466 & 3.746\\
M2 & 220.2 & 99.6\% & 0.826 & 0.691 & 0.938 & 4.378 & 3.994\\
M3 & 254.7 & 98.3\% & 0.906 & 0.809 & 0.974 & 4.438 & 4.576\\
M4 & 321.6 & 94.5\% & 0.890 & 0.787 & 0.830 & 3.996 & 4.162\\
M5 & 1437.0 & 44.0\% & 0.909 & 0.779 & 0.886 & 4.161 & 4.283\\
M6 & 665.7 & 98.7\% & 0.901 & 0.810 & 0.982 & 4.619 & 4.633\\
M7 & 917.1 & 90.8\% & 0.894 & 0.805 & 0.974 & 4.647 & 4.614\\
M8 & 1281.3 & 69.4\% & 0.868 & 0.739 & 0.887 & 3.986 & 4.069\\
M9 & 968.9 & 84.1\% & 0.889 & 0.790 & 0.986 & 4.660 & 4.544\\
M10 & 1740.6 & 13.9\% & 0.929 & 0.750 & 1.000 & 4.700 & 4.400\\
\bottomrule
\end{tabular}
\end{table}

\FloatBarrier

\section{Prompts and Input Templates}
\label{app:prompts}

All prompts and input templates used in VAO-Bench construction and evaluation are provided in the Supplementary Material. This includes the complete templates for source-artifact generation, instruction and annotation generation, target-artifact generation, native and agent-based solving, explicit-instruction and reconstruction ablations, and automated judging.

The Supplementary Material also provides the visual-quality and nine domain-validity rubrics, together with the corresponding input placeholders and rendering specifications. We provide these materials in full to support reproducibility while keeping the main appendix focused on the benchmark construction and evaluation procedures.

\end{document}